\documentclass[acmsmall,screen,nonacm]{acmart}
\AtBeginDocument{%
  \providecommand\BibTeX{{%
    \normalfont B\kern-0.5em{\scshape i\kern-0.25em b}\kern-0.8em\TeX}}}

\setcopyright{acmlicensed}
\acmJournal{CSUR}
\acmYear{2024} \acmVolume{1} \acmNumber{1} \acmArticle{1} \acmMonth{1}\acmDOI{10.1145/3654662}

\acmJournal{CSUR}

\usepackage{subfigure}
\usepackage{supertabular}
\usepackage{tabu}
\usepackage{booktabs,tabularx} %
\usepackage{multirow}

\usepackage{makecell}

\DeclareDocumentCommand{\dataset}{m}{\textsc{#1}}

\begin{document}

\title{A Survey on Responsible LLMs: Inherent Risk, Malicious Use, and Mitigation Strategy
}

	\author{Huandong Wang}
        \orcid{0000-0002-6382-0861}
	\affiliation{%
		\institution{Department of Electronic Engineering, Tsinghua University}
		\city{Beijing}
		\country{China}}
	\email{wanghuandong@tsinghua.edu.cn}
\authornote{Equal contributions.}

	\author{Wenjie~Fu}
\orcid{0009-0004-9879-946X}
	\affiliation{%
		\institution{Huazhong University of Science and Technology}
		\country{China}}
\email{wjfu99@outlook.com}
	\authornotemark[1]

	\author{Yingzhou~Tang}
\orcid{0009-0007-6927-245X}
	\affiliation{%
		\institution{Department of Electronic Engineering, Tsinghua University}
		\city{Beijing}
		\country{China}}
\email{tyz23@mails.tsinghua.edu.cn}
	\authornotemark[1]

	\author{Zhilong~Chen}
    \orcid{0000-0002-2692-5429}
	\affiliation{%
		\institution{Department of Electronic Engineering, Tsinghua University}
		\city{Beijing}
		\country{China}}
	\email{czl20@mails.tsinghua.edu.cn}
	\authornotemark[1]

	\author{Yuxi~Huang}
\orcid{0009-0007-1182-125X}
	\affiliation{%
		\institution{Huazhong University of Science and Technology}
		\country{China}}
\email{huangyuxi@hust.edu.cn}
	\authornotemark[1]

		\author{Jinghua~Piao}
        \orcid{0000-0003-2256-4256}
	\affiliation{%
		\institution{Department of Electronic Engineering, Tsinghua University}
		\city{Beijing}
		\country{China}}
	\email{pjh22@mails.tsinghua.edu.cn}
	\authornotemark[1]

		\author{Chen~Gao}
        \orcid{0000-0002-7561-5646}
	\affiliation{%
		\institution{BNRist, Tsinghua University}
		\city{Beijing}
		\country{China}}
	\email{chgao96@gmail.com}

		\author{Fengli~Xu}
        \orcid{0000-0002-5720-4026}
	\affiliation{%
		\institution{Department of Electronic Engineering, Tsinghua University}
		\city{Beijing}
		\country{China}}
	\email{fenglixu@tsinghua.edu.cn}

		\author{Tao~Jiang}
        \orcid{0000-0002-8482-1046}
	\affiliation{%
		\institution{Huazhong University of Science and Technology}
		\country{China}}
        \email{taojiang@hust.edu.cn}

		\author{Yong Li}
        \orcid{0000-0001-5617-1659}
	\affiliation{%
		\institution{Department of Electronic Engineering, Tsinghua University}
		\city{Beijing}
		\country{China}}
        \email{liyong07@tsinghua.edu.cn}

\begin{abstract}

While large language models (LLMs) present significant potential for supporting numerous real-world applications and delivering positive social impacts, they still face significant challenges in terms of the \emph{inherent risk} of privacy leakage, hallucinated outputs, and value misalignment,
and can be \emph{maliciously used} for generating toxic content and unethical purposes after been jailbroken.
Therefore, in this survey, we present a comprehensive review of recent advancements aimed at mitigating these issues, organized across the four phases of LLM development and usage: data collecting and pre-training, fine-tuning and alignment, prompting and reasoning, and post-processing and auditing.
We elaborate on the recent advances for enhancing the performance of LLMs in terms of privacy protection, hallucination reduction, value alignment, toxicity elimination, and jailbreak defenses.
In contrast to previous surveys that focus on a single dimension of responsible LLMs, this survey presents a unified framework that encompasses these diverse dimensions, providing a comprehensive view of enhancing LLMs to better serve real-world applications.
\end{abstract}

\renewcommand{\shortauthors}{Wang et al.}

\maketitle

\section{Introduction}


The superior abilities of large language models (LLMs) in natural language processing have brought us one step closer to realizing artificial general intelligence (AGI).
Language is not merely a medium for communication~\cite{xi2023rise}, but also plays a fundamental role in the development of thought and higher-level cognitive processes~\cite{vygotsky2012thought}.
Therefore, the breakthrough in language processing has considerably enhanced AI's capacities in both external social communication and internal cognitive functions, advancing its potential to tackle real-world challenges~\cite{tomavsev2020ai}.
Consequently, LLMs have enabled a wide spectrum of applications from increasing social productivity to promoting social good, such as
code generation~\cite{liu2024your},
autonomous agent construction~\cite{xi2023rise,gao2023large}, 
urban planning~\cite{zhou2024large},
improving the efficiency of clinical~\cite{thirunavukarasu2023large} and education~\cite{kasneci2023chatgpt},
assisting disadvantaged groups~\cite{seo2024maidr}, paving the way toward a good society where human and AI coexist~\cite{tomavsev2020ai,roberts2021achieving}.

However, while we enjoy the convenience provided by LLMs, we have to recognize the potential dangers and vulnerabilities they may pose.
Overlooking them has already led to numerous painful lessons~\cite{el2023man,mauran2023whoops}.
For example, it is reported that Samsung employees unintentionally shared confidential source code and meeting recordings with ChatGPT during tasks like error checking, code optimization, and transcription, exposing sensitive business information of the company~\cite{el2023man}.
Another famous example is the ``Grandma Exploit'', where malicious adversaries can persuade LLMs through role-playing as a grandma to elicit responses that LLMs would typically restrict, e.g., how to create a bomb or write source codes for harmful malware~\cite{grandma1,grandma2,zhou2024quantifying}, which poses severe risks to society.
In a more tragic case, a man tragically ended his life after AI chatbots encouraged him to commit suicide during a six-week conversation about the climate crisis, worsening his eco-anxiety and suicidal thoughts~\cite{mauran2023whoops}.
Therefore, it is important to build responsible LLMs, which are capable of mitigating these potential risks and vulnerabilities and preventing such tragedies.

Inspired by these lessons, numerous researchers begin to pay attention to constructing responsible LLMs on different aspects~\cite{das2024security,yan2024protecting,neel2023privacy,tonmoy2024comprehensive,chakraborty2024detoxbench,wu2024fine}.
Among them, the most thoroughly explored aspect is aligning LLMs with human values~\cite{schramowski2022large,choenni2024self,lin2023unlocking,rao2023ethical,sun2024principle,sun2024principle,lee2023rlaif,bai2022constitutional,wu2024fine,ouyang2022training,korbak2023pretraining,lu2022quark,solaiman2021process}.
These works mainly utilize LLM fine-tuning techniques, and incorporate the value dimensions given by representative human value theories, e.g., the Schwartz Theory of Basic Values, and the Moral Foundations Theory~\cite{yao2024clave,schwartz2012overview}.
Nevertheless, responsible LLMs cannot be accomplished merely through value alignment techniques.
For example, hallucinated outputs do not inherently misalign with human values, but significantly reduce LLMs' usability. 
Consequently, multiple aspects should be considered cohesively to construct responsible LLMs.
Simultaneously, the dangers and vulnerabilities of LLMs can arise not only from the generated text but also within their internal mechanisms. For example, existing studies have revealed that LLM logits for specific sentences can expose private information~\cite{mattern2023membership,fu2023practical,fu2024miatuner}. 
Therefore, it is crucial to consider potential dangers and vulnerabilities throughout the entire lifecycle of LLMs, and employ mitigation strategies at the corresponding phase of developing and utilizing LLMs to build responsible LLMs.

\begin{figure}
  \centering
  \includegraphics[width=1.0\linewidth]{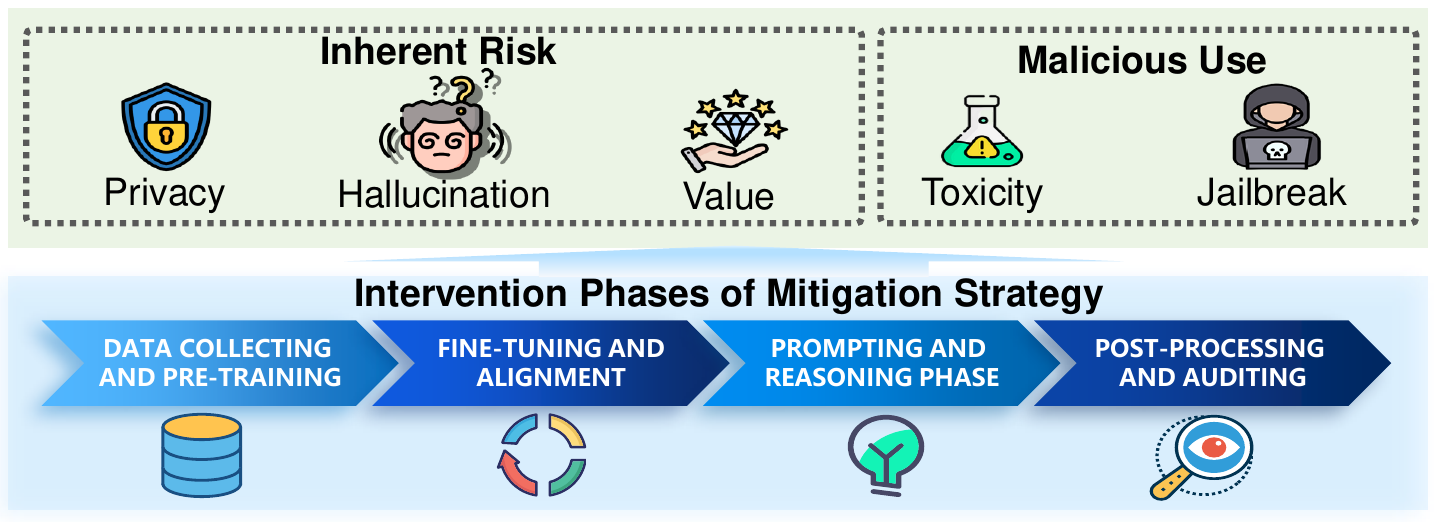}
  \caption{Framework of this survey. Five major dimensions of LLM responsibility are involved, divided into inherent risk (privacy, hallucination, value), and malicious use (toxicity and jailbreak). In terms of the mitigation strategies, we divide the whole process of developing and utilizing LLMs into four intervention phases.}\label{fig:framework}
  \vspace{-0.3cm}
\end{figure}

Overall, we categorize the potential dangers and vulnerabilities of LLMs into two categories, i.e., inherent risk and malicious use (see Figure~\ref{fig:framework}).
In terms of \textbf{inherent risk}, 
LLMs could potentially reveal sensitive information from their utilized corpora for pre-training or fine-tuning, thereby raising issues of \textit{privacy} leakage~\cite{das2024security,yan2024protecting,neel2023privacy}.
Meanwhile, it is well-known that LLMs may experience \textit{hallucinations}, resulting in the production of texts that are inaccurate and misleading~\cite{tonmoy2024comprehensive}.
Finally, since the values embedded in LLM-generated texts usually directly reflect the distribution of their training data, often sourced from the Internet, there exists a substantial risk that LLMs will overfit to a narrow set of human values or even amplify undesirable and unethical content, resulting in~\textit{value} mismatch.
In terms of \textbf{malicious use}, LLMs could be utilized to produce content with \textit{toxicity}, such as hate speech, harassment,
cyberbullying, causing harm to humans~\cite{chakraborty2024detoxbench}.
In addition, malicious users may \textit{jailbreak} LLMs to bypass their safety constraints for fraudulent purposes~\cite{xu2024comprehensive,liu2024autodan}.
In terms of \textbf{mitigation strategies}, we divide the whole process of developing and utilizing LLMs into four phases, i.e., \emph{data collecting and pre-training}, \emph{fine-tuning and alignment phase}, \emph{prompting and reasoning}, and \emph{post-processing and auditing phase}, and classify existing studies based on the phase they operate on.

In this paper, we systematically summarize the recent studies toward constructing responsible LLMs, covering techniques enhancing LLMs in terms of privacy protection, hallucination reduction, value alignment, toxicity elimination, and jailbreak defenses. 
Compared with existing surveys, which mainly focus on a single dimension for improving the responsibility of LLM (e.g., privacy~\cite{das2024security,yan2024protecting,neel2023privacy} or  hallucination~\cite{tonmoy2024comprehensive}), our survey employs a unified framework to analyze and compare existing studies covering a broader range of dimensions.
This comprehensive analysis helps us understand the commonalities and differences across these different dimensions, facilitating a deep understanding of the current state and future directions of constructing responsible LLMs.

The structure of this survey is as follows. We begin by reviewing related surveys regarding achieving responsible LLMs and compare their major difference with our paper in Section~\ref{sec:relatedreviews}. Then, we formally introduce and compare the four phases composing the whole process of preparing and utilizing LLMs in specific applications in Section~\ref{sec:4phases}. Afterward, we detailedly introduce existing studies about achieving responsible LLMs in terms of privacy protection, hallucination reduction, value alignment, toxicity elimination, and jailbreak defense from Section~\ref{sec:privacy} to Section~\ref{sec:jailbreak}, respectively.
Finally, we summarize open challenges and future directions for responsible LLMs in Section~\ref{sec:future} and conclude our paper in Section~\ref{sec:conclusion}.

\section{Related Reviews}\label{sec:relatedreviews}

Reviews related to our paper are summarized in Table~\ref{Tab:existingsurvey}.
Specifically,
\citeauthor{yao2023instructions}~\cite{yao2023instructions} surveyed alignment goals of large language models, which include human instructions, human preferences, and human values. Further, they investigated the definition of different alignment goals and benchmarks and methods to evaluate these alignment goals.
\citeauthor{wang2023aligning}~\cite{wang2023aligning}  summarized alignment technologies with human expectations in terms of three aspects, i.e., data collection, training methodologies, and model evaluation.
\citeauthor{yao2024survey}~\cite{yao2024survey}  reviewed related papers regarding security and privacy. Specifically, they mainly discussed existing works in terms of three aspects, i.e.,  LLM's positive impacts on security and privacy, LLM's negative impacts on security and privacy, and vulnerabilities and defenses in LLMs.
\citeauthor{das2024security}~\cite{das2024security} investigated the vulnerabilities in terms of security attacks and privacy attacks, and also reviewed defense mechanisms against security and privacy attacks.
\citeauthor{yan2024protecting}~\cite{yan2024protecting}  focused on the privacy concerns of LLM. Specifically, they revealed the privacy risk of LLMs in terms of passive privacy leakage and active privacy attacks, and summarized major privacy protection mechanisms against passive privacy leakage and active privacy attacks. 
\citeauthor{neel2023privacy}~\cite{neel2023privacy}  comprehensively investigated the privacy issues regarding LLMs, which include memorization, privacy attacks, privacy-preserving techniques, and copyright.
\citeauthor{huang2023survey}~\cite{huang2023survey}  investigated the cause of hallucination in LLMs, as well as detection methods, benchmarks, and mitigation methods of hallucination in LLMs.
\citeauthor{chowdhury2024breaking}~\cite{chowdhury2024breaking} reviewed thirty-two techniques addressing hallucination in LLMs, presenting a taxonomy based on dataset utilization, common tasks, feedback mechanisms, and
retriever types, and highlighting methods like RAG and CoNLI. 
\citeauthor{yao2023instructions}~\cite{yao2023instructions} 
reviewed existing works in terms of alignment goals, which are then categorized into three levels, including human instructions, human preferences, and human values. 
\citeauthor{yi2024jailbreak}~\cite{yi2024jailbreak} proposed a taxonomy of jailbreak attack and defense methods for LLMs, categorizing attacks into black-box and white-box types and defenses into prompt-level and model-level strategies, further detailing sub-classes and evaluation methods to inspire secure LLM development.
\citeauthor{chowdhury2024breaking}~\cite{chowdhury2024breaking} presented a comprehensive survey on the security vulnerabilities of LLMs, focusing on adversarial attacks, data poisoning, and privacy risks. They evaluated attack methodologies, model resilience, and defense strategies, providing insights into LLM integrity and user trust.
As we can observe, these existing surveys mainly focus on a single dimension for constructing responsible LLMs. 
However, different aspects of LLM responsibility are deeply interconnected. For example, value alignment and privacy protection share a common principle of preventing the disclosure of harmful information. At the same time, conflicts also exist between dimensions~\cite{zhang2023defending,yao2023value}.
Consequently, it is crucial to incorporate different dimensions of LLM responsibility comprehensively.
Different from them, our survey employs a unified framework to analyze and compare existing studies covering a broader range of dimensions. 
Furthermore, we also provide a unique perspective of categorizing and comparing existing approaches based on the phases of LLM development and utilization on which they operate.

\begin{table*}[t]
\renewcommand{\arraystretch}{1.00}
  \begin{center}
\small
\resizebox{\linewidth}{!}{%
  \begin{tabular}{|c|c|p{12.2cm}<{\centering}|}
  \hline
   \textbf{Paper} & \textbf{Dimension}  & \textbf{Content}\\ \hline
   \cite{yan2024protecting} & Privacy & Passive
privacy leakage, active privacy attacks, and major privacy protection
mechanisms. \\ \hline
   \cite{neel2023privacy} & Privacy & Memorization,
privacy attacks, privacy-preserving techniques, and copyright.\\ \hline
   \cite{yao2024survey} & Security and privacy & LLM's positive impacts on security and privacy, negative impacts on security and privacy, and vulnerabilities and defenses.\\ \hline
   \cite{das2024security} & Security and privacy & Security attacks, privacy attacks, and defense mechanism.\\ \hline
   \cite{huang2023survey} & Hallucination & Detection methods, benchmarks, and mitigation methods. \\ \hline
   \cite{tonmoy2024comprehensive} & Hallucination  & Hallucination mitigation techniques and a taxonomy based on their dataset utilization, tasks, feedback mechanisms, and
retriever types. \\ \hline
      \cite{yao2023instructions} & Value & Definition of value alignment goal, evaluation benchmarks and methods.\\ \hline
   \cite{wang2023aligning} & Value & Data collection, training methodologies, and model evaluation\\ \hline
   \cite{yao2023instructions}  & Value &  Three level of alignment goals including human instructions, human preferences, and human values.\\ \hline
   \cite{yi2024jailbreak} & Jailbreak & Jailbreak attack and defense methods for LLMs, including black-box and white-box attacks, prompt-level and model-level defense strategies.\\  \hline
   \cite{chowdhury2024breaking} & Jailbreak and privacy & Adversarial attacks, data poisoning, and privacy risks, and evaluation in terms of attack effectiveness, model resilience, and defense strategies.\\  \hline
  \end{tabular}
  }
  \end{center}
    \caption{Related reviews to responsible LLMs.}
  \label{Tab:existingsurvey}
  \vspace{-0.5cm}
\end{table*}


\section{Four phases for Employing Mitigation Strategy towards Responsible LLMs}\label{sec:4phases}


\subsection{Data collecting and pre-training phase}

The data collecting and pre-training phase is the first phase in the LLM life cycle, where the LLM is trained on large corpora of text data to learn the language patterns and structures~\cite{devlin2019BERTPretra,radfor2018Improving,touvro2023LLaMAOpen,du2022GLMGenera}.
In this phase, multiple sources of data, including text data from books, academic materials, encyclopedia, code, social media, and webpages, are collected to ensure the LLM can learn from various domains~\cite{liu2024DatasetsL}.
However, for the data that is collected from the internet and other untrusted sources, it is crucial to ensure the data is clean and free from noise, bias, and other unwanted information~\cite{zhao2023SurveyLar,galleg2024BiasFairn}.
Therefore, quality filtering and privacy reduction processes are necessary to handle the data before the pre-training phase.
For example, \citeauthor{chen2024DataJuicer} proposed a data cleaning method called Data-Juicer, which provides a systematic way to process and clean the data for pre-training~\cite{chen2024DataJuicer}.
\citeauthor{lauren2022BigScience} employed a rule-based method to remove personally identifiable information (PII) from the data~\cite{lauren2022BigScience}.
These methods can help to improve the quality of the data, which is essential for the performance and sanity of the LLM~\cite{longpr2024Pretrainer}, and also to protect the privacy of the users and the data subjects from privacy leakage and other potential risks~\cite{carlin2022Quantifyin,miresh2022Quantifyin}.
However, both the quantity and quality of the data are important for the pre-training phase.
\citeauthor{longpr2024Pretrainer} discovered that when toxic content is filtered out from the pre-training data, the LLM will generate less toxic content in the downstream tasks, while reducing the performance on some other tasks~\cite{longpr2024Pretrainer}.

\subsection{Fine-tuning and alignment phase}

Taking place directly after the data collecting and pre-training phase, the fine-tuning and alignment phase aims to adapt the generic knowledge of pre-trained models to some target tasks~\cite{radford2018improving}, e.g., mitigating unwanted outputs and boosting the performances on specific downstream tasks~\cite{achiam2023gpt}. For example, a representative and widely-adopted method for fine-tune large language models is Reinforcement Learning from Human Feedback (RLHF)~\cite{ouyang2022training}. Taking a pre-trained language model as a start, it adopts supervised methods to learn from human instruction data and is subsequently optimized according to human annotators' comparison-based preferences through reinforcement learning. As such, LLMs can better reflect human preferences and act more like humans.

The presence of the fine-tuning and alignment phase makes it possible to make the full use of the large quantity of data and eases the efforts in model training. Sometimes the quality of the pre-training corpora may not be that satisfying and the corpora may contain unwished contents, e.g., biases~\cite{gallegos2024bias} and toxicity~\cite{he2024you}. Removing them in the data collection and pre-training phase may help, but could be too costly when the size of the corpora explodes. In these circumstances, conducting adequate fine-tuning and alignment turns to be an effective and viable solution: it not only retains the information available in the corpora, but also saves the exhaustive efforts in the complicated manipulation of the original data, e.g., data filtering~\cite{gallegos2024bias}. Moreover, with fine-tuning and alignment, practitioners can take the full advantage of the knowledge inherited in the existing pre-trained models and no longer need to train everything from scratch~\cite{min2023recent}. This significantly alleviates their burden in training while maintaining promising results on the downstream tasks of focus.

\subsection{Prompting and reasoning phase}

The prompting and reasoning phase is an important aspect of utilizing LLMs to undertake various challenging tasks~\cite{zhao2023survey,minaee2024large}. Numerous studies have pointed out that the quality of prompts largely determines the reasoning capability of LLMs~\cite{marvin2023prompt,sahoo2024systematic}. For example,
a vague prompt like ``Write about the company's goals'' might generate a generic response, lacking in detail and relevance. However, a more precise prompt, such as ``Write a two-paragraph introduction for a business proposal that outlines our company's goals for expanding into the renewable energy market in 2024,'' would lead the LLMs to generate a far more tailored and useful response. Therefore, to better utilize LLMs’ capabilities in reasoning, some researchers have developed various prompting methods~\cite{huang2023towards,chang2024survey}. One well-known approach is Chain-of-Thought (CoT) prompting, which allows LLMs to generate intermediate reasoning steps before arriving at a final answer, enhancing their problem-solving and decision-making accuracy~\cite{wei2022emergent}. Based on the idea of CoT, a variety of variants have been proposed. For example, the Zero-shot-CoT~\cite{kojima2022large} is designed to trigger the reasoning capability by incorporating prompts like ``Let's think step by step''. Recently, inspired by heuristic searching, Yao et al.~\cite{yao2024tree} design the Tree-of-Thought prompting method. Moreover, Yu et al.~\cite{yu2023thought} propose the Thought Propagation method by leveraging analogous problems and solutions to further enhance the quality of reasoning.

Although numerous studies have highlighted the superior reasoning capability of LLMs~\cite{huang2023towards,chang2024survey}, the prompting and reasoning phase is vulnerable, especially when facing inappropriate prompts and deliberate attacks~\cite{xie2023defending,gallegos2024bias}. On the other hand, improving this phase can further contribute to building responsible LLMs~\cite{xie2023defending,gallegos2024bias}. For example, Xie et al.~\cite{xie2023defending} find LLMs could be jailbroken by specific adversarial prompts. However, they also point out that self-reminder prompting, a design imposed on the promoting and reasoning phase, significantly reduces the success rate of jailbreak attacks~\cite{xie2023defending}. Overall, the prompting and reasoning phase plays a pivotal role in building responsible LLMs.

\subsection{Post-processing and auditing phase}

The post-processing and auditing phase is the last feasible phase in the LLM life cycle, following the completion and stabilization of the preceding three phases. During this phase, the LLM is thoroughly pre-trained and does not permit additional fine-tuning or alignment, regardless of whether these processes have already been carried out. Additionally, the prompting and reasoning templates are meticulously curated to meet the specific task requirements and are not subject to revision. Thus, LLM only provides a pure black-box access API to output generated texts based on given queries. Audit-and-Process is the general pipeline to improve the responsibility of LLMs in this phase, where auditing and processing algorithms are introduced to detect and obliterate the latent harmful information in LLMs' generated texts~\cite{lukas2023analyzing, kim2024propile, jin2024jailbreakzoo, xu2024rejection}. This pipeline has been adopted by several studies to prevent LLM from directly releasing text that contains personally identifiable information~\cite{lukas2023analyzing, kim2024propile}, abusive language~\cite{wang2020detect, song2023measuring, bodapati2019neural, baratalipour2020abusive} and cyberbullying comments~\cite{cheng2024leveraging, jin2024guard, kumar2024bias, li2024semantic}. Existing auditing algorithms can be divided into rule-based~\cite{sekine2004definition, krupka2005description, chiticariu2010domain} and machine learning-based approaches~\cite{lukas2023analyzing, lample2016neural, phan2020collective, tran2017named}. Early rule-based methods are simply and straightforward for deploying, which leads to better explainability of errors~\cite{chiticariu2010domain}, but lack flexibility and accuracy~\cite{nadeau2007survey, chiticariu2010domain}. Thus, modern methods who have achieved state-of-the-art results most often resort to machine learning techniques~\cite{lukas2023analyzing, lample2016neural, phan2020collective, tran2017named}. After auditing generated text, there are three common safeguard strategies to obiliterage detected harmful information: scrub~\cite{lukas2023analyzing}, regeneration~\cite{jin2024jailbreakzoo, watts2024voterbased} and rejection~\cite{xu2024rejection}. For cases where only certain tokens or chunks within the generated sentence are harmful, such as personally identifiable information~\cite{lukas2023analyzing}, the post-processing algorithm will mask or replace those elements~\cite{lukas2023analyzing}. For situations where the overall generated content is harmful, such as racism and sexism statements~\cite{haim2024what}, the post-processing algorithm will consider regenerating a harmless text~\cite{jin2024jailbreakzoo} or directly refusing to answer prompts that are suggestive and harmful~\cite{xu2024rejection}.
Compared to earlier phases, deploying strategies during the auditing and post-processing phases offers greater compatibility, as this strategy only requires pure black-box access to the LLM, making it suitable for all categories of LLMs. Moreover, this pipeline does not involve any adjustments to the fine-tuned LLM, making it more promising in terms of reducing performance decay in the LLM. However, the Audit-and-Process pipeline may significantly increase the time and computational cost required during the inference phase, as regeneration can be time-consuming and may require multiple iterations to bypass the auditing algorithm~\cite{jin2024jailbreakzoo}. The auditing algorithms in this pipeline may also affect the continuous, real-time, or interactive generation of LLMs, as some algorithms might require waiting for the LLM to finish generating~\cite{lukas2023analyzing, lample2016neural, phan2020collective, tran2017named, sekine2004definition, krupka2005description, chiticariu2010domain}.

\newcommand{\wjfublod}[1]{{\vspace{2pt} \bf \noindent #1 \hspace{0.5pt}}}
\definecolor{mygray}{rgb}{0.784,0.784,0.784}
\newcommand*{\greysquare}{\textcolor{mygray}{\blacksquare}}

\section{Privacy}\label{sec:privacy}
\begin{figure}[t]
\vspace{-0.2cm}
    \centering
    \includegraphics[width=0.93\linewidth]{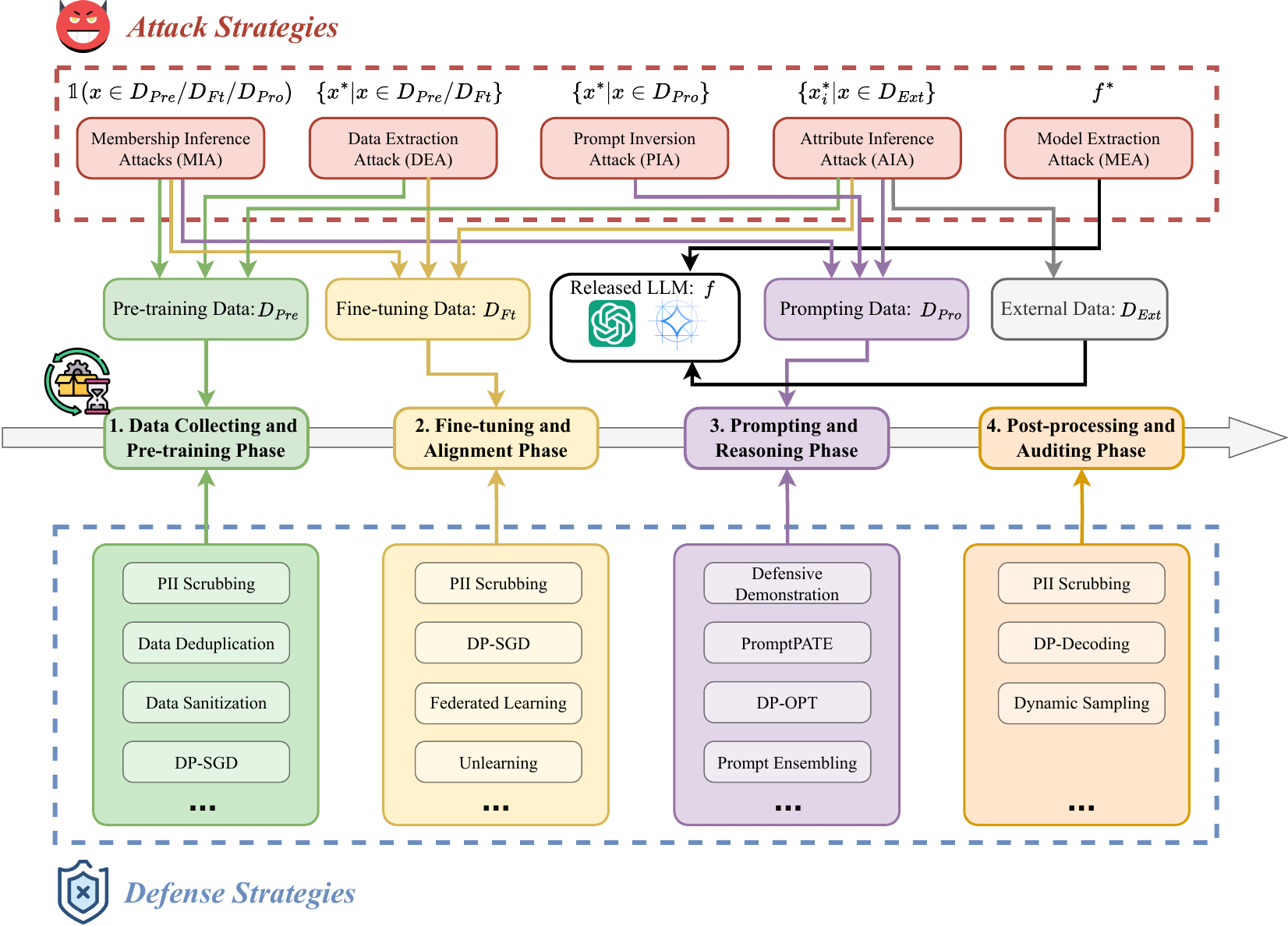}
\vspace{-0.2cm}
    \caption{Overview of the attack and defense methods for evaluating and improving LLMs in terms of privacy.}
    \label{fig:privacy}
\vspace{-0.4cm}
  \end{figure}

LLMs are typically pre-trained on massive and myriad corpora fetched from websites, code repositories, user posts, and other sources that may contain a large amount of privacy-sensitive information~\cite{lukas2023analyzing}. Due to the strong few-shot capabilities of LLMs and the development of prompting and reasoning techniques, some privacy data, such as clinical diagnosis notes, appears as demonstrations in the In-Context Learning (ICL) scenario~\cite{wang2023prompt}. Therefore, privacy risks associated with LLMs persist throughout their entire life cycle~\cite{peris2023privacy}, making it crucial to assess and mitigate these risks to build responsible LLMs.
In this section, we comprehensively examine privacy risks throughout the four phases in the life cycle of LLMs. Initially, we explore the privacy vulnerability of LLM and the adversarial methods developed to reveal or evaluate the privacy risk in LLMs (\S \ref{par: privacy risk}). Subsequently, we delve into the safeguards and countermeasures proposed to mitigate these privacy risks (\S \ref{par: privacy defense}).  Furthermore, we provide a high-level overview of the typical attack and defense strategies in Figure~\ref{fig:privacy} as the outline of the following content.

\subsection{Privacy Risks in LLMs}\label{par: privacy risk}

In this section, we embark on an inclusive review of the typical privacy risks associated with LLMs. We will examine specific attack methods and corresponding privacy risks associated with privacy-sensitive data appearing at different phases of the LLM lifecycle. Encompassing data from LLM training corpus data to prompting demonstration data and even external data that does not belong to LLM.
Except for exposing privacy-sensitive data, the undisclosed parameters of LLM are also shown to be vulnerable to specific attacks~\cite{carlini2024stealing}. Various privacy evaluation benchmarks are proposed to assess the privacy risks of LLMs, as depicted in Table~\ref{tab:privacy_benchmark}. These benchmarks are designed to evaluate the privacy risks of LLMs from different attack dimensions, including (1) \textbf{Membership Inference Attack (MIA)}~\cite{shokri2017membership}: inferring whether a given text record is used for training LLM, (2) \textbf{Data Extraction Attack (DEA)}~\cite{carlini2021extracting, carlini2023extracting}: extracting the text records that exist in the training dataset, (3) \textbf{Prompt Inversion Attack (PIA)}~\cite{morris2023language}: stealing the private prompting texts, and (4) \textbf{Attribute Inference Attack (AIA)}~\cite{wang2023decodingtrust, staab2023memorization,lukas2023analyzing}: deducing the private or sensitive information from training texts, prompting texts or external texts. \textbf{Model Extraction Attack (MEA)}~\cite{carlini2024stealing}, replicating the parameters of the LLM, is rarely explored and without a dedicated benchmark. 
As depicted in Table~\ref{tab:privacy_attack}, we additionally consolidate these attack techniques based on the disclosed private information, the attack dimension, the level of model access necessitated by the attacker, as well as the dataset or benchmark utilized, and the specific target model on which they are assessed. 
In the following content, we will delineate five distinct types of privacy breaches, encompassing pre-training data, fine-tuning data, prompting data, external data, and model parameters.

\textbf{Pre-training Corpus Data.} A significant portion of the privacy risk in the training set is attributed to the memorization traces left by the training data on the LLM, which allow specific attacks to infer relevant features of the training data. These memorization-based attacks can be categorized into Membership Inference Attacks~\cite{shokri2017membership, carlini2022membership, ye2022enhanced} and Data Extraction Attacks~\cite{fredrikson2015model}. Membership inference attacks on pre-training corpus data are referred to as the pre-training data detection task, aiming to ascertain whether a specific textual sample was encountered by LLM during its pre-training phase~\cite{fu2024miatuner}. Because of the extensive scale of pre-training corpora, the opacity of the training data distribution, and the reduced number of training epochs, detecting pre-training samples is non-trivial~\cite{zhang2024mink} and has only garnered limited attention. \citeauthor{shi2023detecting}~\cite{shi2023detecting}  propose Min-K\% that takes average over the $k$ minimum likelihoods of tokens for detection. \citeauthor{zhang2024mink}~\cite{zhang2024mink} further advance Min-K\%, incorporate with the insight that training data tends to be around the local maximum in the probability distribution~\cite{fu2023practical}. In contrast to curate a meticulous MIA metric, \citeauthor{fu2024miatuner}~\cite{fu2024miatuner} suggest to instruct LLMs themselves to play as a pre-training data detector and introduce the MIA-Tuner. This method manifests higher confidence and feasibility on both aligned and unaligned LLMs. Data extraction attacks are a more advanced category of privacy attacks that intend to extract partial content deeply memorized by LLM. \citeauthor{carlini2021extracting}~\cite{carlini2021extracting} first introduce an untargeted training data extraction method, in which the adversary generates enormous samples and then filters the content in the pre-training set. 
The subsequent works mostly focus on targeted data extraction that recovers the suffix text given the prefix~\cite{carlini2022quantifying}. \citeauthor{carlini2022quantifying}~\cite{carlini2022quantifying} adopt the targeted extraction as a tool to evaluate the memorization of LLM. \citeauthor{yu2023bag}~\cite{yu2023bag} further investigate and benchmark several overlooked tricks, such as Top-$k$, Temperature, and Nucleus-$\eta$~\cite{holtzman2019curious}, for improving data extraction. \citeauthor{zhang2023ethicist}~\cite{zhang2023ethicist} and \citeauthor{ozdayi2023controlling}~\cite{ozdayi2023controlling} share a similar idea that triggers LLM to emit more pre-training data through prompt-tuning~\cite{lester2021power}.

\textbf{Fine-tuning Corpus Data.} Since both pre-training and fine-tuning data are essentially training data for LLMs, most privacy attack methods targeting pre-training data are also applicable to fine-tuning data. Furthermore, fine-tuning datasets are typically smaller in scale and security-critical since it is usually encompassed with more private downstream data. Therefore, some existing works more concentrate on compromising the privacy of fine-tuning data. For example, \citeauthor{mireshghallah2022empirical}~\cite{mireshghallah2022empirical} propose LiRA, an MIA method, for evaluating memorization in fine-tuned Autoregressive Language Models. \citeauthor{mattern2023membership}~\cite{mattern2023membership} design a neighborhood comparison method to detect fine-tuning data with the more practical assumption. \citeauthor{fu2023practical}~\cite{fu2023practical} achieve a better trade-off between attack performance and assumptions in MIAs against fine-tuned LLMs by fine-tuning a reference model based on the data extracted from the target fine-tuned LLM. \citeauthor{wen2024privacy}~\cite{wen2024privacy} introduce a novel privacy backdoor by releasing a tampered pre-trained LLM, which aims to amplify the vulnerability of fine-tuning data to membership inference attack. Similarly, \citeauthor{feng2024privacy}~\cite{feng2024privacy} also employ privacy backdoor to poison pre-trained LLM, which enables adversaries to reconstruct individual fine-tuning data.

\begin{table}[t]
\small
\renewcommand{\arraystretch}{1.2}
\caption{Benchmark for evaluating the risk of privacy leakage of LLMs.}\label{tab:privacy_benchmark}
\vspace{-1em}
\resizebox{\linewidth}{!}{%
\begin{tabular}{|c|p{6cm}|c|c|} 
\hline
\textbf{Benchmark} & \textbf{Information}                                                                                         & \textbf{\textbf{Dimension}} & \textbf{Format}                           \\ 
\hline
WIKIMIA~\cite{shi2023detecting}           & Pre-training data detection benchmark from Wikipedia.                                             & MIA                         & Plain text                                \\ 
\hline
WIKIMIA-24~\cite{fu2024miatuner}        & Updated version of WIKIMIA for evaluating recent LLMs.                     & MIA                         & Plain text, Q-A pair                      \\ 
\hline
MIMIR~\cite{duan2024membership}              & Pre-training data detection benchmark based on the Pile dataset. dataset.                                         & MIA                         & Plain text                                \\ 
\hline
LM-Extraction~\cite{2024googleresearch}     & Data extraction benchmark using easy-to-extract samples from the Pile dataset.             & DEA                         & Prefix-Suffix pair                        \\ 
\hline
LLM-PBE~\cite{li2024llmpbe}            & Privacy evaluation benchmark covering the entire LLM lifecycle. & DEA, MIA, PIA               & Plain text, Q-A pair,~Prefix-Suffix pair  \\
\hline
DECODINGTRUST~\cite{wang2023decodingtrust}            & Trustworthiness benchmark for GPT models from eight perspectives. & DEA, AIA               & Prefix-Suffix pair, Q-A pair, Prompt text  \\
\hline
Instructions-2M~\cite{morris2023language}            & Meta-dataset with 2.33M instructions including prompts for various tasks. & PIA               & Prompt text  \\
\hline
\end{tabular}
\renewcommand{\arraystretch}{1}
}
\vspace{-1em}
\end{table}

\begin{table}[h]
  \small
  \setlength\tabcolsep{5pt}
  \renewcommand{\arraystretch}{1.20}
  \caption{Taxonomy of methods for evaluating the risk of privacy leakage of LLMs in terms of private information, attack dimension, attacker's access, dataset, and target models. \textbf{MIA}: Membership Inference; \textbf{DEA}: Data Extraction Attack; \textbf{PIA}: Prompt Inversion Attack; \textbf{AIA}: Attribute Inference Attack; \textbf{MEA}: Model Extraction Attack; $\square$: White-box access; $\greysquare$: Grey-box access; $\blacksquare$: Black-box access.}\label{tab:privacy_attack}
  \vspace{-1em}
  \resizebox{\linewidth}{!}{%
  \begin{tabular}{|c|c|c|c|c|c|c|} 
  \hline
  \textbf{Paper}                      & \textbf{Private Information}                     & \textbf{Dimension}           & \textbf{Access} & \textbf{Methodology}    & \textbf{Dataset/Benchmark}                           & \textbf{Target Models}                                                              \\ 
  \hline
  \cite{li2023mope}                   & Pre-training Data                                & MIA                          & $\square$       & MoPe                    & Pile                                                 & Pythia                                                                              \\ 
  \hline
  \cite{shi2023detecting}             & Pre-training Data                                & MIA                          & $\greysquare$   & MIN-K\%                 & WIKIMIA                                              & \makecell[c]{Pythia, GPT-NeoX, \\LLaMA, OPT}                                        \\ 
  \hline
  \cite{zhang2024mink}                & Pre-training Data                                & MIA                          & $\greysquare$   & MIN-K\%++               & WIKIMIA, MIMIR                                       & Mamba, Pythia, LLaMA                                                                \\ 
  \hline
  \cite{xie2024recall}                & Pre-training Data                                & MIA                          & $\greysquare$   & ReCall                  & WIKIMIA, MIMIR                                       & \makecell[c]{Pythia, GPT-NeoX, Mamba, \\LLaMA, OPT}                                 \\ 
  \hline
  \cite{fu2024miatuner}               & Pre-training Data                                & MIA                          & $\greysquare$   & MIA-Tuner               & WIKIMIA, WIKIMIA-24                                  & \makecell[c]{Pythia, Falcon, LLaMA, \\LLaMA-2, Mistral, Gemma}                      \\ 
  \hline
  \cite{wang2024conrecall}            & Pre-training Data                                & MIA                          & $\greysquare$   & Con-ReCall              & WIKIMIA, MIMIR                                       & \makecell[c]{Pythia, GPT-NeoX, \\Mamba, LLaMA}                                      \\ 
  \hline
  \cite{mireshghallah2022quantifying} & Pre-training Data                                & MIA                          & $\greysquare$   & LiRA                    & MIMIC-III, i2b2                                      & ClinicalBERT                                                                        \\ 
  \hline
  \cite{carlini2021extracting}        & Pre-training Data                                & DEA                          & $\blacksquare$  & Untargeted Extraction   & Internet Search                                      & GPT-2                                                                               \\ 
  \hline
  \cite{carlini2021extracting}        & Pre-training Data                                & MIA, DEA                     & $\greysquare$   & Zlib                    & Internet Search                                      & GPT-2                                                                               \\ 
  \hline
  \cite{carlini2021extracting}        & Pre-training Data                                & MIA, DEA                     & $\greysquare$   & Lowercase               & Internet Search                                      & GPT-2                                                                               \\ 
  \hline
  \cite{carlini2021extracting}        & Pre-training Data                                & MIA, DEA                     & $\greysquare$   & Smaller ref             & Internet Search                                      & GPT-2                                                                               \\ 
  \hline
  \cite{nasr2023scalable}             & Pre-training Data                                & DEA                          & $\blacksquare$  & Divergence Attack       & AUXDATASET                                           & GPT-3.5-turbo                                                                       \\ 
  \hline
  \cite{yu2023bag}                    & Pre-training Data                                & DEA      & $\blacksquare$  & Tricks                  & LM-Extraction                                        & GPT-Neo                                                                             \\ 
  \hline
  \cite{tang2023assessing}            & Pre-training Data                                & MIA      & $\blacksquare$  & Text Similarity         & SAMsum, CN-NDM, MIMIC                                & BART-base, FLAN-T5                                                                  \\ 
  \hline
  \cite{lukas2023analyzing}           & Fine-tuning Data                                 & DEA, AIA & $\blacksquare$  & PII Extraction           & ECHR, Enron, Yelp-Health                             & GPT-2                                                                               \\ 
  \hline
  \cite{ozdayi2023controlling}        & Pre-training Data                                & DEA                          & $\greysquare$   & Controllable Extraction & LM-Extraction                                        & GPT-Neo, GPT-2                                                                      \\ 
  \hline
  \cite{zhang2023ethicist}            & Pre-training Data                                & DEA                          & $\greysquare$   & ETHICIST                & LM-Extraction                                        & GPT-Neo                                                                             \\ 
  \hline
  \cite{fu2023practical}              & Fine-tuning Data                                 & MIA                          & $\greysquare$   & SPV-MIA                 & AG News, Wikitext, Xsum                              & \makecell[c]{GPT-2, GPT-J, \\LLaMA, Falcon}                                         \\ 
  \hline
  \cite{mattern2023membership}        & Fine-tuning Data                                 & MIA                          & $\greysquare$   & Neighbor Attack         & AG News, Twitter, Wikitext                           & GPT-2                                                                               \\ 
  \hline
  \cite{mireshghallah2022empirical}   & Fine-tuning Data                                 & MIA                          & $\greysquare$   & LiRA                    & \makecell[c]{Wikitext, Enron email,\\ Penn Treebank} & GPT-2                                                                               \\ 
  \hline
  \cite{duan2023privacy}              & \makecell[c]{Prompting Data, \\Fine-tuning Data} & MIA                          & $\greysquare$   & Confidence              & AG News, cb, sst2, rte                               & GPT-2                                                                               \\ 
  \hline
  \cite{wang2023decodingtrust}        & \makecell[c]{Pre-training, \\Prompting Data}     & DEA, AIA                     & $\blacksquare$  & DECODINGTRUST           & DECODINGTRUST                                        & GPT3.5, GPT-4                                                                       \\ 
  \hline
  \cite{morris2023language}           & Prompting Data                                   & PIA                          & $\blacksquare$  & Prompt Inversion        & Instructions-2M                                      & LLaMA-2                                                                             \\ 
  \hline
  \cite{staab2023memorization}        & External Data                                    & AIA                          & $\blacksquare$  & Privacy Inference       & PAN,~PersonalReddit                                  & \makecell[c]{LLaMA, GPT-3.5, PaLM-2,\\ LLaMA-2, GPT-4, Claude-2,\\ Claude-Instant}  \\ 
  \hline
  \cite{carlini2024stealing}          & Model Parameters                                 & MEA                          & $\blacksquare$  & Projection Extraction   & N/A                                                  & \makecell[c]{OpenAI's ada and babbage \\ GPT-2, Pythia, LLaMA}                      \\
  \hline
  \end{tabular}
  }
  \renewcommand{\arraystretch}{1}
\end{table}

\textbf{Prompting Demonstration Data.} 
The seminal work of in-context-learning (ICL) has revolutionized the field of LLMs~\cite{brown2020language}, which enables LLM adaptation for specific downstream tasks by prompting the model with a series of task demonstrations but without tuning any model's parameters~\cite{dong2022survey}.
Although the prompting demonstration data is not included for training LLMs, its private information can be inferred during the inference phase. Existing studies have demonstrated that LLMs may verbatim regurgitate prompt data in their answers with the appropriate user instruction~\cite{priyanshu2023chatbots, duan2024flocks, wang2023decodingtrust}. Thus, ~\citeauthor{morris2023language}~\cite{morris2023language} formally propose a prompt inversion attack to extract the private prompting text from the LLM. Imagine a healthcare application where private clinical diagnosis notes are used as task demonstrations to instruct LLM to assist in diagnosis and treatment. However, a malicious query may be tampered with to steal these private demonstrations, which will raise critical privacy risks. 

\textbf{External Textual Data.} In addition to the internal data, the external textual data from an unrelated domain or source fed into LLMs can also lead to considerable privacy risks. For example, \citeauthor{staab2023memorization}~\cite{staab2023memorization} conduct the first exploration of LLM's analytical ability to infer personal private information, such as address, income, and sex, from texts. They demonstrate that by scraping the content of a user’s online posts and feeding them to an LLM, malicious agents can infer private and sensitive information never intended to be disclosed by the users. They verified that LLMs can act like private detectives, tracing subtle clues in users' posts to extract their attributes from seemingly non-private text, e.g., \textit{``there is this nasty intersection on my commute, I always get stuck there waiting for a hook turn''}, from which LLM can deduce the user's address is Melbourne, since \textit{``a hook turn''} is a traffic maneuver particularly used in Melbourne. Their experiments depict that LLMs can achieve up to 85\% accuracy in inferring personal attributes with only a fraction of the cost (100$\times$) and time (240$\times$) required by humans.

\textbf{Model Parameters.} Privacy risks in general machine learning models are not only limited to the data but also include the parameters of models~\cite{tramer2016stealing}. Model stealing attacks targets at duplicating (i.e., ``stealing'') the victim ML models, where the duplicated model $f'$ should perform equivalent to the target model $f$ on the whole input space. For any valid input $x$, $f'(x) \approx f(x)$ is expected. Model stealing attacks have been well investigated in traditional ML models and revealed substantial privacy risks~\cite{milli2019model, jagielski2020high}.
However, only a minority of studies focus on extracting specific properties of LLMs~\cite{zanella2021grey,zanella2021grey,zanella2021grey}, as this task is considerably more challenging for LLMs compared to typical ML models, primarily due to their orders-of-magnitude larger parameter scale. Besides, the existing stealing attacks designed for LLMs aims to recover more limited knowledge, or make a stronger assumption.
\citeauthor{wei2020leaky}~\cite{wei2020leaky} enable the attacker on the same server as the LLM to infer the sizes of hidden layers. \citeauthor{zanella2021grey}~\cite{zanella2021grey} propose a method to extract the final layer, while assuming the target model is composed of a public pre-trained encoder and a private final layer. \citeauthor{carlini2024stealing}~\cite{carlini2024stealing} first successively and precisely extract nontrivial information from production LLMs (e.g., ChatGPT) in a pure black-box setting. Particularly, the proposed methods can extract the entire embedding projection layer of a transformer model. Besides, the proposed method has been verified can extract the projection layer of OpenAI's ada and babbage language models with less than \$20 budget.





\begin{table}[h]
\small
\setlength\tabcolsep{5pt}
\renewcommand{\arraystretch}{1.2}
\caption{Taxonomy of methods for improving the privacy protection performance of LLMs in terms of the defended private information, the attack dimension that can be defended against, as well as the dataset or benchmark, and the target model on which they are evaluated. \textbf{MIA}: Membership Inference; \textbf{DEA}: Data Extraction Attack; \textbf{PIA}: Prompt Inversion Attack; \textbf{AIA}: Attribute Inference Attack; \textbf{MEA}: Model Extraction Attack; \textbf{DP}: Data Collecting and pre-training phase; \textbf{FA}: Fine-tuning and alignment phase; \textbf{PR}: Prompting and reasoning phase; \textbf{PA}: Post-processing and auditing phase.}\label{tab:privacy_defense}
\vspace{-1em}
\resizebox{\linewidth}{!}{%
\begin{tabular}{|c|c|c|c|c|c|c|} 
\hline
\textbf{Paper}                         & \textbf{Phase}       & \textbf{Private Information}                        & \textbf{Dimension}   & \textbf{Mitigation Method}                                                  & \textbf{Dataset/Benchmark}                    & \textbf{Target Models}                                                              \\ 
\hline
\cite{lee2022deduplicating}            & DP                   & Pre-training Data                                   & MIA, DEA             & Data Deduplication                                                    & \makecell[c]{C4, RealNews, \\LM1B, Wiki40B}   & T5                                                                                  \\ 
\hline
\cite{kandpal2022deduplicating}        & DP                   & Pre-training Data                                   & MIA, DEA             & Data Deduplication                                                    & OpenWebText, C4                               & GPT-2                                                                               \\ 
\hline
\cite{li2021largea}                    & DP; FA               & \makecell[c]{Pre-training Data, \\Fine-tuning Data} & MIA, DEA, AIA        & DP-SGD                                                                & \makecell[c]{GLUE, E2E, DART, \\Persona-Chat} & GPT-2, BERT, RoBERTa                                                                \\ 
\hline
\cite{anil2022largescale}              & DP                   & \makecell[c]{Pre-training Data}                     & MIA, DEA, AIA        & DP-SGD                                                                & \makecell[c]{Wikipedia, \\BooksCorpus}        & BERT                                                                                \\ 
\hline
\cite{wu2023depn}                      & DP; FA               & \makecell[c]{Pre-training Data, \\Fine-tuning Data} & MIA, DEA, AIA        & DEPN                                                                  & Enron                                         & BERT                                                                                \\ 
\hline
\cite{lukas2023analyzing}              & DP; FA                   & Fine-tuning Data                                    & DEA, AIA             & PII Scrubbing                                                         & \makecell[c]{ECHR, Enron, \\Yelp-Health}      & GPT-2                                                                               \\ 
\hline
\cite{basu2022benchmarking}            & FA                   & Fine-tuning Data                                    & MIA, DEA, AIA        & DP-FL                                                                 & \makecell[c]{Depression, \\Sexual Harrasment} & \makecell[c]{BERT, RoBERTa, \\DistillBERT, ALBERT}                                  \\ 
\hline
\cite{hoory2021learning}               & FA                   & Fine-tuning Data                                    & MIA, DEA, AIA        & DP-SGD                                                                & MIMIC-III                                     & BERT                                                                                \\ 
\hline
\cite{yu2021differentially}            & FA                   & Fine-tuning Data                                    & MIA, DEA, AIA        & DP-Tune                                                               & DART                                          & GPT-2                                                                               \\ 
\hline
\cite{yu2021large}                     & FA                   & Fine-tuning Data                                    & MIA, DEA, AIA        & RGP                                                                   & \makecell[c]{MNLI, QQP, \\QNLI, SST-2}        & BERT                                                                                \\ 
\hline
\cite{mireshghallah2022differentially} & FA                   & Fine-tuning Data                                    & MIA, DEA, AIA        & DPKD, DPIMP                                                           & \makecell[c]{MNLI, QQP, \\QNLI, SST-2}        & BERT                                                                                \\ 
\hline
\cite{duan2023flocks}                  & FA                   & Fine-tuning Data                                    & MIA, DEA             & PromptDPSGD                                                           & GLUE                                          & RoBERTa                                                                             \\ 
\hline
\cite{ozdayi2023controlling}           & FA                   & Pre-training Data                                   & DEA                  & Controllable Extraction                                               & LM-Extraction                                 & GPT-Neo, GPT-2                                                                      \\ 
\hline
\cite{fu2024miatuner}                  & FA                   & Pre-training Data                                   & MIA                  & MIA-Tuner                                                             & \makecell[c]{WIKIMIA, \\WIKIMIA-24}           & \makecell[c]{Pythia, Falcon, LLaMA, \\LLaMA-2, Mistral, Gemma}                      \\ 
\hline
\cite{chen2023unlearn}                 & FA                   & Pre-training Data                                   & MIA, DEA, AIA        & Unlearning                                                            & IMDB, SAMSum                                  & T5                                                                                  \\ 
\hline
\cite{jang2023knowledge}               & FA                   & Pre-training Data                                   & MIA, DEA, AIA        & Unlearning                                                            & Pile                                          & GPT-Neo, OPT                                                                        \\ 
\hline
\cite{eldan2023whos}                   & FA                   & Pre-training Data                                   & MIA, DEA, AIA        & \makecell[c]{Approximate \\Unlearning}                                & Harry Potter                                  & LLaMA                                                                               \\ 
\hline
\cite{kassem2023preserving}            & FA                   & Pre-training Data                                   & MIA, DEA, AIA        & DeMem                                                                 & Pile                                          & GPT-2,~GPT-Neo                                                                      \\ 
\hline
\cite{wang2023decodingtrust}           & PR                   & \makecell[c]{Prompting Data}                        & AIA                  & \makecell[c]{Defensive\\Demonstrations}                     & DECODINGTRUST                                 & GPT3.5, GPT-4                                                                       \\ 
\hline
\cite{duan2023privacy}                 & PR                   & \makecell[c]{Prompting Data, \\Fine-tuning Data}    & MIA                  & Prompt Ensembling                                                     & \makecell[c]{AG News, cb, \\SST-2, RTE}       & GPT-2                                                                               \\ 
\hline
\cite{duan2023flocks}                  & PR                   & Prompting Data                                      & MIA, PIA             & PromptPATE                                                            & \makecell[c]{DBPedia, SST-2, \\AG News, TREC} & GPT-3                                                                               \\ 
\hline
\cite{hong2023dpopt}                   & PR                   & Prompting Data                                      & MIA, PIA             & DP-OPT                                                                & \makecell[c]{SST-2, TREC, \\Mpqa, Disaster}   & Vicuna, LLaMA-2, GPT-3                                                              \\ 
\hline
\cite{wu2023privacypreserving}         & PR                   & Prompting Data                                      & MIA, PIA             & \makecell[c]{Noisy Vote, \\ESA, KSA}                                  & \makecell[c]{PFL-DocVQ, \\SAMSum}             & LLaMA, GPT-3                                                                        \\ 
\hline
\cite{tang2023privacypreserving}       & PR                   & Prompting Data                                      & MIA, PIA             & \makecell[c]{DP Few-shot \\Generation}                                & \makecell[c]{AG News, TREC, \\DBPedia}        & GPT-3                                                                               \\ 
\hline
\cite{majmudar2022differentially}      & PA                   & Pre-training Data                                   & MIA, DEA, AIA        & DP-Decoding                                                           & \makecell[c]{Common Crawl,\\Wikipedia, mC4}   & RoBERTa-style                                                                       \\ 
\hline
\cite{ginart2022submix}                & PA                   & Fine-tuning Data                                     & DEA, AIA             & SUBMIX                                                                & \makecell[c]{Wikitext-103, \\BigPatent-G}     & GPT-2                                                                               \\ 
\hline
\cite{staab2023memorization}           & FA;~PA               & External Data                                       & AIA                  & \makecell[c]{PII Removing, \\Better Alignment}                        & PAN,~PersonalReddit                           & \makecell[c]{LLaMA, GPT-3.5, PaLM-2,\\ LLaMA-2, GPT-4, Claude-2,\\ Claude-Instant}  \\ 
\hline
\cite{morris2023language}              & PA                   & Prompting Data                                      & DEA                  & Dynamic Sampling                                                      & Instructions-2M                               & LLaMA-2                                                                             \\ 
\hline
\cite{carlini2024stealing}             & DP; PA               & Model Parameters                                    & MEA                  & \makecell[c]{API Limitation, \\Noise Addition, \\Architectural Amend} & N/A                                           & \makecell[c]{OpenAI's ada and babbage, \\GPT-2, Pythia, LLaMA}                      \\ 
\hline
\multicolumn{1}{c}{}                   & \multicolumn{1}{c}{} & \multicolumn{1}{c}{}                                & \multicolumn{1}{c}{} & \multicolumn{1}{c}{}                                                  & \multicolumn{1}{c}{}                          & \multicolumn{1}{c}{}                                                               
\end{tabular}
}
\renewcommand{\arraystretch}{1}
\end{table}

\subsection{Privacy Protection for LLMs}\label{par: privacy defense} 

In this section, we conduct a detailed review of the defense strategies proposed to address the aforementioned privacy risks. The existing defense methods throughout the whole lifecycle of LLM, including the four phases from the data collecting and pre-training phase to the post-processing and auditing phase that we have mentioned in Section~\ref{sec:4phases}. We will categorize all defense methods into four classes based on the phases at which they intervene in the deployment and usage of LLMs. As illustrated in Table~\ref{tab:privacy_defense}, we further summarize these defense methods in terms of the defended private information, the attack dimension that can be defended against, as well as the dataset or benchmark, and the target model on which they are evaluated.

\textbf{Data Collecting and pre-training phase.} In the upstream phase of LLM deployment, injecting defensive strategies during the data collection and pre-training phase is most likely to fundamentally eliminate or mitigate the privacy risks of LLM. Several studies suggest to conduct data sanitization for improving privacy before pre-training~\cite{lee2022deduplicating, kandpal2022deduplicating, lukas2023analyzing}. For instance, \citeauthor{lee2022deduplicating}~\cite{lee2022deduplicating} observe that most document-level deduplicated web-scraped datasets still have large-scale sentence-level duplication. To tackle this issue, they propose efficient sentence-level deduplication methods and decrease the training consumption by approximately 10 times. \citeauthor{kandpal2022deduplicating}~\cite{kandpal2022deduplicating} then demonstrate that the sentence-level duplication can mitigate the privacy risks caused by the model memorization, including MIA, DEA, and AIA. Except from the data deduplication, some works also explore to remove or replace all Personally Identifiable Information (PII) tagged by Named Entity Recognition (NER) modules~\cite{lukas2023analyzing}. Other studies have attempted to incorporate differential privacy (DP) into the pre-training phase~\cite{li2021largea, anil2022largescale, wu2023depn}, which can provide a rigorous privacy guarantee and has a commendable post-processing property (i.e., any subsequent computation or transformation performed will not increase the risk of exposing individual data points~\cite{dwork2014algorithmic}). Specifically, \citeauthor{li2021largea}~\cite{li2021largea} propose a ghost clipping technique to address the computational challenge and enormous memory usage of employing DP-SGD on LLM and verify that LLMs can be strong DP-learner. \citeauthor{anil2022largescale}~\cite{anil2022largescale} train BERT with DP to achieve high accuracy by proposing mega-batches, which scales up the training batch size to millions. Furthermore, \citeauthor{wu2023depn}~\cite{wu2023depn} present a method to detect and edit private neurons in pre-trained LLMs to address privacy risks, which can reduce the model memorization of private data.

\textbf{Fine-tuning and alignment phase.} Compared with pre-training data, fine-tuning data is usually more vulnerable to adversaries due to some specific characteristics, such as smaller scale and higher confidentiality~\cite{yu2021differentially}. Thus, some works focus on developing a privacy-preserving fine-tuning pipeline to guarantee the private information in fine-tuning data will not be disclosed after the fine-tuned LLMs are released~\cite{basu2022benchmarking, hoory2021learning, yu2021differentially, yu2021large}. 
\citeauthor{hoory2021learning}~\cite{hoory2021learning} first propose to fine-tune a differentially private BERT model with small performance degradation through a novel word-piece algorithm and DP-SGD. \citeauthor{yu2021differentially}~\cite{yu2021differentially} introduce a meta-framework for achieving a better tradeoff between differential privacy and utility in fine-tuning autoregressive language models, such as GPT-2. 
\citeauthor{yu2021large}~\cite{yu2021large} propose the reparametrized gradient perturbation (RGP) for applying DP on fine-tuning large models, such as large vision and large language models. \citeauthor{duan2023flocks}~\cite{duan2023flocks} present the PromptDPSGD algorithm to conduct private gradient descent on the soft prompt embeddings prepended to the LLM’s private input.
Besides, fine-tuning is more efficient in computational consumption than pre-training and can avoid training the LLM from scratch~\cite{han2024parameterefficient}. Consequently, some other works consider patching during the fine-tuning stage to fix privacy leaks caused during the pre-training phase~\cite{ozdayi2023controlling, fu2024miatuner, jang2023knowledge}. Such as \citeauthor{ozdayi2023controlling}~\cite{ozdayi2023controlling} adopt prompt-tuning to control the vulnerability of LLM against extraction attack, which significantly decreases the extraction rate with competitive PPL values. 
\citeauthor{fu2024miatuner}~\cite{fu2024miatuner} introduce two defense strategies with their proposed MIA-Tuner framework to fine-tune both aligned and unaligned LLMs to defend against MIA. Recently, the privacy-preserving regulations, including the EU’s General Data Protection Regulation (GDPR) and US’s California Consumer Privacy Act (CCPA), have also required that \textit{users have the right to be forgotten}. As a consequence, several studies have started to investigate how to fine-tune a pre-trained LLM to unlearn specific content that has included pre-training corpora and needs to be forgotten~\cite{chen2023unlearn, jang2023knowledge, eldan2023whos, kassem2023preserving}. \citeauthor{jang2023knowledge}~\cite{jang2023knowledge} simply fine-tune LLM on the text to be forgotten while negating (maximizing) the original loss function. 
\citeauthor{chen2023unlearn}~\cite{chen2023unlearn} introduce to fine-tune a lightweight unlearning layer via a selective teacher-student formulation, and they further propose a fusion mechanism to combine different unlearning layers to handle a sequence of forgetting operations. \citeauthor{eldan2023whos}~\cite{eldan2023whos} point out that simply negating the loss function for unlearning may not yield satisfying results. In contrast to fine-tuning LLM on the text that needs to be forgotten, they suggest to replace the idiosyncratic expressions in the target data with generic counterparts, and then fine-tuning LLMs on the regenerated text.

\textbf{Prompting and reasoning phase.} The widespread use of in-context learning in downstream tasks under LLM has highlighted the risk of prompt demonstration data being stolen. To address this issue, some studies have proposed privacy-preserving in-context learning, which ensures that private demonstration examples are not disclosed while maintaining the performance of LLM few-shot learning~\cite{duan2023privacy, duan2023flocks, hong2023dpopt, wu2023privacypreserving, tang2023privacypreserving}. \citeauthor{duan2023privacy}~\cite{duan2023privacy} introduce to ensemble multiple prompted LLMs with disjoint demonstrations to aggregate the prediction probability vectors, which will moderately mitigate the privacy risks in a certain extent. \citeauthor{duan2023flocks}~\cite{duan2023flocks} further propose PromptPATE to orchestrate a noisy vote among an ensemble of LLMs presented with different demonstrations, which follows the general flow of standard PATE (private aggregation of teacher ensembles)~\cite{papernot2018scalable} and first strike rigorous differential privacy in ICL. 
However, PromptPATE~\cite{duan2023flocks} requires to transfer the knowledge from private labeled data to an unlabeled public dataset which may not exist in practice.
\citeauthor{hong2023dpopt}~\cite{hong2023dpopt} believe that LLM services may be hosted by untrustworthy providers, and directly sending sensitive private information to the provider can be dangerous. Thus, they introduce DP-OPT to fine-tune a differentially-private prompt in a local trusted LLM and then transfer this prompt to the cloud LLM. The aforementioned privacy-preserving ICL methods are all stuck in classification task~\cite{duan2023flocks, hong2023dpopt}, which will limit the generalizability for employing the defense methods on some ICL tasks. \citeauthor{wu2023privacypreserving}~\cite{wu2023privacypreserving} propose DP-ICL, a general paradigm for privatizing ICL in text classification and language generation tasks, which generates differentially private responses through a noisy aggregation among an ensemble of LLM’s responses based on disjoint exemplar sets. Similarly, \citeauthor{tang2023privacypreserving}~\cite{tang2023privacypreserving} also present a comparable concept regarding the construction of LLM ensembles. However, in contrast to the aggregation procedure in DP-ICL~\cite{wu2023privacypreserving}, they no longer combine the output results of each prompted LLM. Instead, they predict the next generated token by consolidating noisy probabilities, which may be more applicable to various ICL tasks and could potentially yield enhanced performance.

\textbf{Post-processing and auditing phase.} In numerous practical scenarios, the defender may not have the access to intervene in the deployment of LLMs in earlier phases. For instance, certain LLM providers solely provide services via black-box APIs. Consequently, integrating audit and post-processing modules during the LLM text generation phase represents a more adaptable approach to bolster LLM privacy. Several studies have explored to defend diverse attack methods during the post-processing and auditing phase~\cite{morris2023language, ginart2022submix, majmudar2022differentially, staab2023memorization, carlini2024stealing}. \citeauthor{majmudar2022differentially}~\cite{majmudar2022differentially} introduce the DP-Decoding, a method to conceal the pre-training data, which perturbs the probability distribution of the next token predicted by the LLM. Although the DP-Decoding provides a possible solution, the performance degradation under a moderate differential privacy is nearly unacceptable. 
\citeauthor{ginart2022submix}~\cite{ginart2022submix} aim to protect the fine-tuning data through a differentially private decoding mechanism named SUBMIX that follows the PATE~\cite{papernot2018scalable} to mix the ensemble predictions on the next token. However, SUBMIX~\cite{ginart2022submix} requires an ensemble of fine-tuned LLM that substantially increases the computation and memory consumption. \citeauthor{morris2023language}~\cite{morris2023language} design a dynamic sampling method to prevent the private information in the demonstration prompt is stolen, which dynamically adjusts the decoding parameters, such as temperature, top-$p$, and top-$k$, during the inference stage.  \citeauthor{staab2023memorization}~\cite{staab2023memorization} propose to defend against attribute inference attacks on external text data by eliminating PII from the output results of LLMs. \citeauthor{carlini2024stealing}~\cite{carlini2024stealing} believe that by adding a sufficient amount of noise to the output logits of any given query, it would be possible to thwart the model extraction attack.

\section{Hallucination}

Despite the rapid advancement of LLMs, hallucinations have emerged as one of the most vital concerns surrounding their use~\cite{farquhar2024detecting,huang2023survey,zhang2023siren,ji2023survey,li2024dawn}. Hallucinations are often referred to as LLMs' generating content that is nonfactual or unfaithful to the provided information~\cite{farquhar2024detecting,huang2023survey,zhang2023siren,ji2023survey}. Therefore, hallucinations can be typically categorized into two main classes. The first is factuality hallucination, which describes the discrepancy between LLMs' generated content and real-world facts. For example, if LLMs mistakenly take Charles Lindbergh as the first person who walked on the moon, it is a factuality hallucination~\cite{huang2023survey}. The second is faithfulness hallucination, which describes the discrepancy between the generated content and the context provided by the user's instructions or input, as well as the internal coherence of the generated content itself. For example, when LLMs perform the summarizing task, they occasionally tamper with some key information by mistakes, which is a faithfulness hallucination. Despite increasing concerns about hallucinations in LLMs, our understanding of their causes and effective mitigation strategies remains largely incomplete. Below we will discuss these two points based on the four important phases of LLMs.

\subsection{Hallucination Risks in LLMs}

\begin{figure}
  \vspace{-0.2cm}
  \centering
  \includegraphics[width=0.9\linewidth]{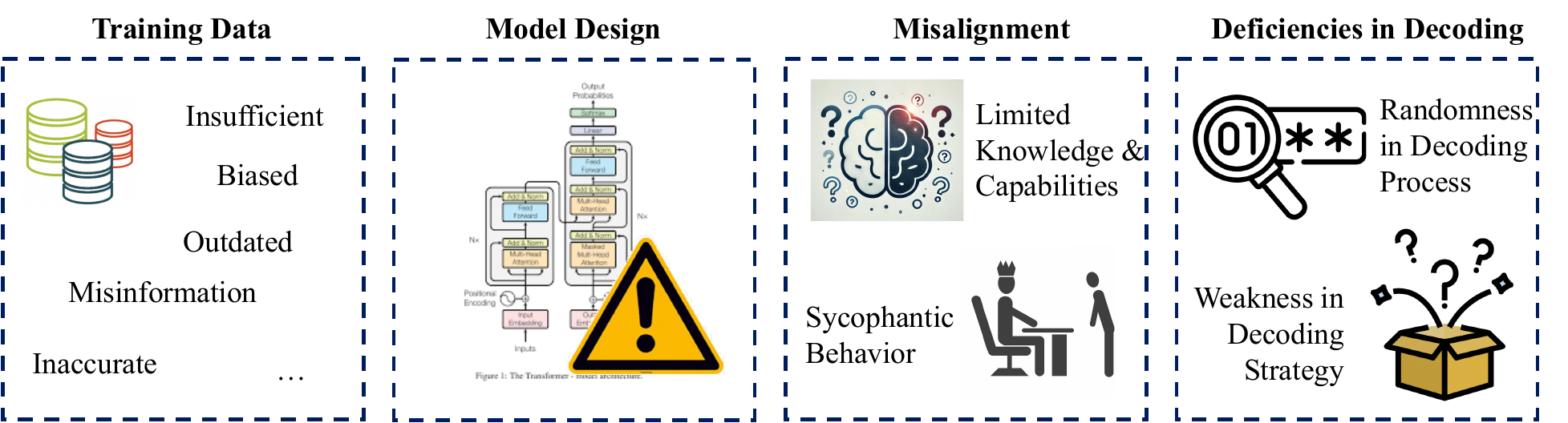}
  \caption{\centering Potential causes of hallucinations in LLMs.}
  \label{fig:hualluci}
  \vspace{-0.2cm}
\end{figure}

Great efforts have been made to explore the origin of hallucinations in LLMs~\cite{farquhar2024detecting,huang2023survey,zhang2023siren,ji2023survey,lin2022truthfulqa,zhang2023language,schulman2023reinforcement,sharma2023towards,aksitov2023characterizing,shi2023trusting,yang2017breaking}. They mainly attribute hallucinations to the insufficient capability of current LLMs in training-related phases, encompassing data collection and pre-training phase, fine-tuning and alignment phase, as well as prompting and reasoning phase. 
After systematically reviewing existing studies investigating hallucinations, we summarize four major causes of hallucinations in LLMs (Figure~\ref{fig:hualluci}), which will be discussed in detail in the following paragraphs.

\textbf{Training Data.} Prior work on data collection and pre-training phase mainly focuses on two main directions. The first direction discusses the insufficiency and poor quality of training data. Indeed, pre-training data determines the knowledge of LLMs~\cite{penedo2023refinedweb,onoe2022entity}, the quality of which is also essential for factual and faithful LLMs. Lin et al.~\cite{lin2022truthfulqa} point out that LLMs are mimicking human falsehoods and propose a benchmark to evaluate whether LLMs can answer truthfully the designed questions. Moreover, besides falsehoods inherent in human languages, the pre-training dataset also contains humans' emphasis on some words or concepts, which leads to duplication bias~\cite{huang2023survey,hernandez2022scaling}. Such duplication limits LLMs to memorizing the input and thereby producing hallucinations~\cite{hernandez2022scaling}. Furthermore, LLMs are also affected by human society-like social biases~\cite{huang2023survey,bang2024measuring,rottger2024political}. These biases result in hallucinations biased toward what is more common in the pre-training dataset~\cite{huang2023survey,bang2024measuring,rottger2024political}. Finally, facts are occasionally time-sensitive, and the pre-training data may not always reflect the latest updates~\cite{penedo2023refinedweb,onoe2022entity}. This also causes the happening of hallucinations in LLMs. 

\textbf{Model Design.} Besides the pre-training data, researchers also point out that the transformer-based architecture is not perfect~\cite{li2023batgpt,liu2024exposing,hahn2020theoretical}. This architecture follows a causal language modeling paradigm and depends on a bi-directional prediction method, which allows it to capture complex unidirectional relations~\cite{li2023batgpt}. This insufficient learning
could potentially increase the risks of hallucinations~\cite {li2023batgpt,huang2023survey}. Moreover, prior works also find that the self-attention module, the core of the transformer-based architecture, suffers from attention glitches~\cite{huang2023survey,liu2024exposing,hahn2020theoretical}.

\textbf{Misalignment.} Misalignment in the fine-tuning and alignment phase also triggers the occurrence of hallucinations in LLMs~\cite{huang2023survey,schulman2023reinforcement,sharma2023towards,rrv2024chaos}. Schulman et al.~\cite{schulman2023reinforcement} have pointed out that LLMs are fine-tuned and aligned to produce content beyond their knowledge and capabilities, which might potentially increase the risks of hallucinations. Moreover, some studies have found that due to misalignment, LLMs can exhibit sycophantic behavior, meaning they tend to choose answers that cater to human preferences, even at the expense of factual accuracy~\cite{sharma2023towards,rrv2024chaos}.

\textbf{Deficiencies in Decoding.} Hallucinations happen also due to shortcomings in the prompting and reasoning phase~\cite{huang2023survey,shi2023trusting,aksitov2023characterizing,yang2017breaking}. Randomness plays an important role in the decoding process of LLMs~\cite{chuang2023dola,dziri2021neural}, which provides diverse generation results. However, the randomness also increases the risk of hallucinations~\cite{chuang2023dola,dziri2021neural}. Moreover, numerous studies have pointed out that inherent weaknesses in the current decoding strategies are responsible for hallucinations~\cite{shi2023trusting,yang2017breaking}. Shi et al.~\cite{shi2023trusting} discover that LLMs always frequently place their attention on nearby words, leading to a strong local focus and a notable weakness in capturing broader contextual information. Yang et al.~\cite{yang2017breaking} demonstrate that the performance of most LLMs that are based on softmax structures is limited by the problem of softmax bottleneck.

\begin{table}[]
  \vspace{-0.2cm}
\renewcommand{\arraystretch}{1.00}
\caption{Taxonomy of main causes and mitigation methods for hallucinations in LLMs. \textbf{TD}: training data, \textbf{MD}: model design, \textbf{MA}: Misalignment, \textbf{DD}: deficiencies in decoding, \textbf{N/A}: Not applicable. }\label{tab:hallucination}
\vspace{-1em}
\resizebox{\linewidth}{!}{%
\begin{tabular}{|p{1cm}|p{2cm}|p{2cm}|p{7cm}|p{3cm}|}
\hline
\textbf{Paper}  & \textbf{Type of hallucinations} & \textbf{Main Causes} & \textbf{Target Models} & \textbf{Mitigation Method} \\
\hline
\cite{lin2022truthfulqa}  & Factual & TD & GPT-3, GPT-Neo/J, GPT-2, and a T5-based model & N/A \\
\hline
\cite{zhang2023language}  & Factual & MD,DD & GPT-3.5, GPT-4, and LLaMA2 & N/A \\
\hline
\cite{sharma2023towards}  & Factual & MA & Claude1.3, Claude-2, GPT-3.5, GPT-4, and LLaMA2 & N/A \\
\hline
\cite{shi2023trusting}  & Faithful & DD & OPT, GPT, LLaMA, and FLAN-T5 & Decoder \\
\hline
\cite{bang2024measuring}  & Faithful & N/A & LLaMA-2, Vicuna, Yi-chat, Falcon-inst, Solar-inst, Mistral-inst, and Jais-chat & N/A \\
\hline
\cite{rottger2024political} & Faithful & N/A & GPT-3.5, GPT-4, Zephyr, LLaMA-2, and Mistral & N/A \\
\hline
\cite{liu2024exposing}  & Factual & MD & Self-trained LLMs & Regularization \\
\hline
\cite{rrv2024chaos}  & Factual & MA & LLaMA-2, Orca-2, Mistral, GPT-3.5 & Prompting and RAG \\
\hline
\cite{chuang2023dola}  & Factual & DD & LLaMA & Decoder \\
\hline
\cite{huang2024opera}  & Faithful & DD & Self-trained LLMs & Decoder \\
\hline
\cite{gao2022rarr}  & Factual & N/A & PaLM & RAG \\
\hline
\cite{yao2022react}  & Factual & N/A & PaLM & Prompting \\
\hline
\cite{li2024dawn}  & Factual & TD, MD, MA, and DD & ChatGPT, Claude-2, Text-Davinci, Vicuna, LLaMA-2, Alpaca & Pre-training, fine-tuning, and inference \\
\hline
\cite{wu2023ragtruth}  & Factual & N/A & GPT-3.5, GPT-4, LLaMA-2, Mistral & RAG \\
\hline
\cite{chen2024benchmarking}  & Factual & N/A & ChatGPT, ChatGLM, ChatGLM2, Vicuna, Qwen, and BELLE & RAG \\
\hline
\cite{farquhar2024detecting}  & Factual & N/A & LLaMA-2, Falcon, and Mistral & N/A \\
\hline
\end{tabular}
}
  \vspace{-0.2cm}
\end{table}

\subsection{Hallucination Reduction for LLMs}

As discussed above, hallucinations are a systematic problem, intricately tied to all phases in the training of LLMs. Therefore, it is challenging to address hallucinations by merely adjusting a single phase of the training process. Currently, mitigation strategies focus more on designing better decoders for LLMs~\cite{chuang2023dola,huang2024opera} and post-processing LLMs' generated content~\cite{farquhar2024detecting,gao2022rarr,yao2022react}. 

\textbf{Prompting and reasoning phase.} In the prompting and reasoning phase, researchers have explored how to design better decoders to alleviate hallucinations~\cite{huang2023survey,chuang2023dola,huang2024opera}. From the interpretability perspective, studies on transformer-based language models have demonstrated that lower-level information exists in the earlier layers while more semantic information exists in the later layers~\cite{tenney2019bert,dai2021knowledge,meng2022locating}. Based on the observation, Chuang et al.~\cite{chuang2023dola} propose a method named Decoding by Contrasting Layers (DoLa), which contrasts logits from earlier and later transformer layers, without requiring external knowledge or fine-tuning. Huang et al.~\cite{huang2024opera} notice the prevalence of hallucinations in multi-modal LLMs. Through empirical investigation, they find that hallucinations in multi-modal LLMs lie in their tendency to generate new tokens based on a few summary tokens, but not all the previous important tokens. Therefore, they propose a decoding method that incorporates an over-trust penalty and retrospection-allocation strategy during beam-search decoding, which significantly mitigates hallucination problems.

\textbf{Post-processing and auditing phase.} Great efforts have been made in the post-processing and auditing phase~\cite{farquhar2024detecting,gao2022rarr,yao2022react}. Different from other mitigation strategies, methods in this phase do not need to change the model architecture or re-train the model parameters. Instead, they mitigate hallucinations with the higher-level capabilities of LLMs, e.g., searching~\cite{gao2022rarr} and reasoning~\cite{yao2022react}. Farquhar et al.~\cite{farquhar2024detecting} adopt LLMs themselves to merge semantically-equivalent answers generated by LLMs, and then utilize semantic entropy to measure the uncertainty of LLMs' answers. However, as discussed in Farquhar et al.~\cite{farquhar2024detecting}, their method focuses only on a subset of hallucinations, i.e., confabulations. However, LLMs occasionally produce false answers with high confidence, making these instances difficult for the proposed approach to detect. Gao et al.~\cite{gao2022rarr} follow a retrieval-augmented generation paradigm~\cite{lewis2020retrieval,jiang2023active,gao2023retrieval} and propose a system that automatically searches attributions for LLMs' generated content and revises incorrect outputs. Indeed, the retrieval-augmented generation paradigm is widely adopted to alleviate hallucinations~\cite{shuster2021retrieval,wu2023ragtruth,chen2024benchmarking,gao2023retrieval,yao2022react}. Yao et al.~\cite{yao2022react} have converted retrieval-augmented generation into LLMs' capabilities of reasoning and acting. Their proposed method enables LLM-empowered agents to generate truthful content based on internal planning and external knowledge.





\section{Value}
\subsection{Value-related Risks in LLMs}

As the general capabilities of LLM-empowered systems improve, the negative consequences and risks induced by these systems also get increasingly alarming accordingly, especially in high-stakes areas~\cite{uk2023,chen2024mismeasure}. Although they may not be intentionally introduced, severe problematic issues related to human values can be raised. Specifically, even before language models become extremely large, pre-trained language models have already exhibited a certain degree of value judgments. For example, 
Schramowski et al.~\cite{schramowski2022large} reveal the existence of the moral direction with the sentence embeddings of moral questions.
However, the distribution of the pre-training corpora may not match exactly with that of the human society~\cite{ferrara2023should} and pieces of knowledge are not guaranteed to be equally learned. As a result, value mismatches may occur. 

In terms of the values demonstrated by LLMs, some researchers such as Scherrer et al.~\cite{scherrer2024evaluating} seek to evaluate the moral beliefs of LLMs through theory-driven tests. It has been shown that in unambiguous scenarios with correct answers (e.g., Should I kill a pedestrian on the road?), most LLMs make the same moral choices as the commonsense. However, in ambiguous scenarios with no commonsense agreements (e.g., Should I tell a white lie?), some models show clear inclinations regardless of the inherent moral ambiguity, where human-aligned models show similar intra-model preferences. As such, moral biases are elicited. Similarly, dimensions such as fairness~\cite{huang2024flames,parrish2022bbq,li2023survey,galleg2024BiasFairn}, safety~\cite{huang2024flames,goldstein2023generative,zhang2023safetybench,sun2023safety}, legality~\cite{huang2024flames}, and offense~\cite{deng2022cold} are attended to and raised caution of.

Circumstances like these will be especially concerning if some populations are put into more disadvantaged positions, which exacerbates the existing social inequality and injustice. For example, Santurkar et al.~\cite{santurkar2023whose} show that LLMs are left-leaning and make communities such as the elderly, Mormon, and widowed underrepresented. Even worse, stereotypes can be even more pronounced and societal prejudice and discrimination may be further triggered~\cite{liu2024prejudice,zuiderveen2018discrimination} and when the cultural representations are biased~\cite{kharchenko2024well}. Therefore, to build responsible LLMs, it is crucial to conduct careful value alignments to mitigate the value-related problems. 

\begin{figure}
  \vspace{-0.2cm}
  \centering
  \includegraphics[width=0.9\linewidth]{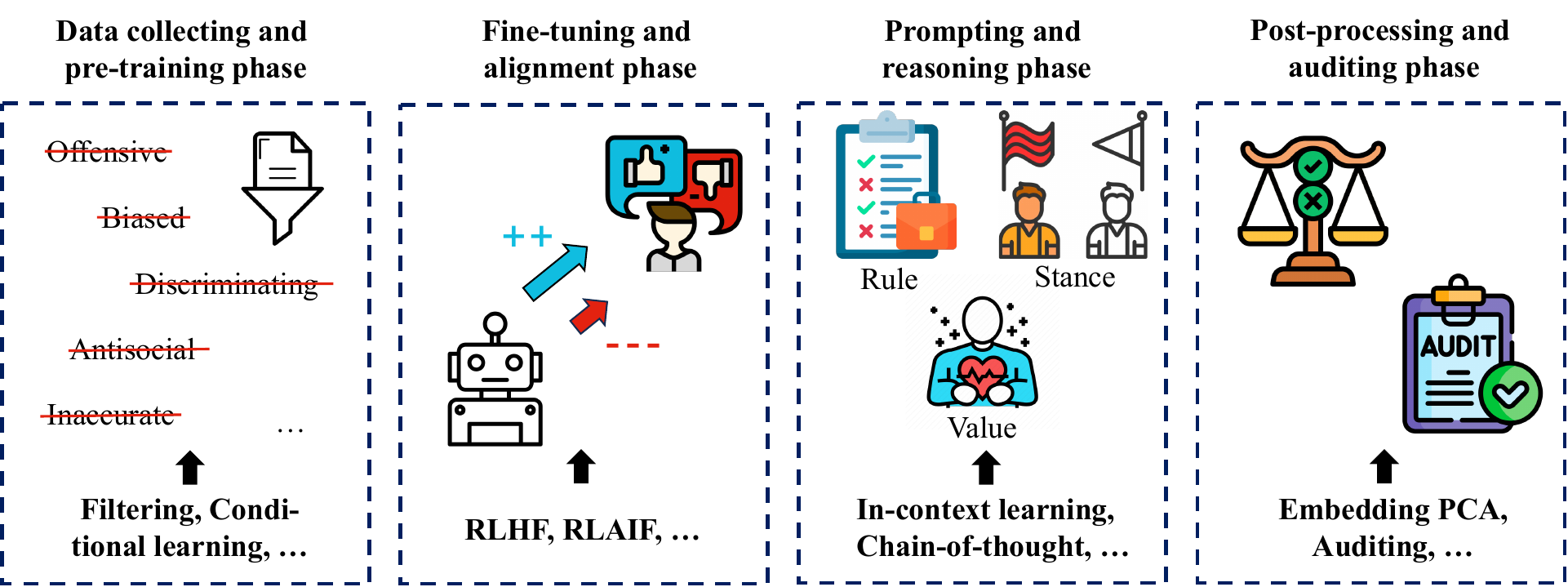}
  \caption{\centering Overview of methods of value alignment for LLMs.}
  \label{fig:value}
  \vspace{-0.2cm}
\end{figure}

\subsection{Value Alignment for LLMs} 

In order to build responsible LLMs, scholars have endeavored to align the values of LLMs with humans, especially in terms of ethicality~\cite{ji2023ai}, to ensure that not only moral principles but also broader social standards of humans are followed so as to better integrate LLMs into society. Here we summarize the representative strategies and provide typical examples for value alignment across the four phases of LLM development and usage (see Figure~\ref{fig:value}).

\textbf{Data collecting and pre-training phase.} Some researchers have endeavored to achieve value alignment in the data collecting and pre-training phase via data manipulation. For example, a first line of researchers consider data filtering~\cite{solaiman2021process}. If we model the training process as conducting maximum likelihood estimation, taking data filtering into consideration can be translated to estimations only on instances whose rewards exceed certain thresholds depicting data qualification. If we do not directly abandon the undesired instances below the thresholds but use token-level unlikelihood by summing up the log-likelihood of all other tokens, we reach unlikelihood training~\cite{welleck2019neural}. Similarly, conditional learning, which prepends texts or segmented tokens with control tokens such as whether they are good, can also take effect~\cite{korbak2023pretraining,lu2022quark}. However, some research warns that these not only do these strategies need an inevitable trade-off between capabilities and value adherence~\cite{korbak2023pretraining}, but they also risk further marginalizing the under-privileged communities and dialects~\cite{xu2021detoxifying,welbl2021challenges}. 

\textbf{Fine-tuning and alignment phase.} Value alignment is most commonly achieved in the fine-tuning and alignment phase, where Reinforcement Learning with Human Feedback (RLHF)~\cite{ouyang2022training} has been the most representative and best-known technique. The procedure of RLHF can be primarily divided into three stages: supervised fine-tuning, reward modeling, and reinforcement-learning-based policy optimization. In supervised fine-tuning, a high-quality dataset on human demonstrations is collected and the pre-trained model is fine-tuned on the collected dataset to have basic mimicry of human behaviors. In reward modeling, annotators' rankings over comparisons of responses are collected and recognized as the reflection of human preferences. Taking these human preferences as references, a reward model is trained. Finally, in reinforcement-learning-based policy optimization, LLMs' generation of different responses from a prompt is regarded as a bandit environment where different responses are assigned varying rewards according to the reward model in the previous step. The parameters of the LLMs are optimized via reinforcement learning to maximize the expected reward on a defined set of prompts for training. In these ways,  human preferences are passed from human annotators to the reward model and then to the LLMs, making the values of the LLMs aligned with humans. 
Some subsequent research proposes variants of the RLHF procedure for value alignment. For example, Wu et al.~\cite{wu2024fine} propose fine-grained RLHF that takes the detailedly annotated category and density of the violating values as signals for sentence-level reward in LLM optimization. Bai et al. demonstrate the possibility of substituting human labels by the combination of a limited number of listed rules plus AI feedback, deriving Reinforcement Learning from AI Feedback (RLAIF)~\cite{bai2022constitutional,lee2023rlaif} that shows the promise to perform comparably well as RLHF. Similarly, some other researchers seek to reduce dependence on human annotation through techniques such as self-alignment~\cite{sun2024principle}.

\textbf{Prompting and reasoning phase.} Value alignment can also be achieved in the prompting and reasoning phase through in-context alignment. This is particularly prominent when value pluralism is present, where multiple value stances are reasonable~\cite{rao2023ethical,sorensen2024roadmap}. For example, facing moral dilemmas, we cannot determine a generic preference and should instead rely on specific moral stances. Rao et al.~\cite{rao2023ethical} achieve this by introducing partially-ordered ethical policies at different levels into the prompts, e.g., from a coarse and general level-2 policy ``justice is more valued than compassion'', to a specific level-0 policy ``[name] prioritizes justice and being impartially treated rather than concerning about his neighbors' cultural beliefs''. In-context learning has also been proven to be effective. As demonstrated by Lin et al.~\cite{lin2023unlocking}, combining in-context learning with structured restyled examples and system prompts for tuning-free alignment can lead to impressive performances across multiple aspects for evaluation. Similarly, Choenni and Shutova~\cite{choenni2024self} proposes to inject cultural-specific values by selecting examples reflecting a culture best for in-context learning.

\begin{table}[]
  \vspace{-0.2cm}
\renewcommand{\arraystretch}{1.00}
\caption{Taxonomy of methods for value alignment of LLMs. \textbf{DP}: data collecting and pre-training phase; \textbf{FA}: fine-tuning and alignment phase; \textbf{PR}: prompting and reasoning phase; \textbf{PA}: post-processing and auditing phase.}\label{tab:value}
\vspace{-1em}
\resizebox{\linewidth}{!}{%
\begin{tabular}{|p{1cm}|p{1cm}|p{4.5cm}|p{3cm}|p{5.5cm}|}
\hline
\textbf{Paper} & \textbf{Phase} & \textbf{Studied/Proposed Model} & \textbf{Approach} & \textbf{Dimension} \\
\hline
\cite{solaiman2021process} & DP & GPT-3 & Filtering & Harm, value targeting \\
\hline
\cite{lu2022quark} & DP & GPT-2 & Conditional training & Harm, negative sentiment, repetition \\
\hline
\cite{korbak2023pretraining} & DP & GPT-2 & Pretraining with HF & Harm, personally identifiable information \\
\hline
\cite{ouyang2022training} & FA & GPT-3, InstructGPT & RLHF & Human preference, truthfulness, harm \\
\hline
\cite{wu2024fine} & FA & T5 & RLHF & Relevance, factuality, information completeness \\
\hline
\cite{bai2022constitutional} & FA & N/A & RLAIF & Helpfulness, harmlessness \\
\hline
\cite{lee2023rlaif} & FA & PaLM 2 & RLAIF & Human preference, harmlessness \\
\hline
\cite{sun2024principle} & FA & LlaMa &  Self alignment & Helpfulness, relevance, accuracy, detail \\
\hline
\cite{rao2023ethical} & PR & GPT-3, ChatGPT, GPT-4 & ICL & Virtue, deontology, consequentialism \\
\hline
\cite{lin2023unlocking} & PR & LlaMA-2, Mistral & ICL & Helpfulness, clarity, factuality, depth, engagement, safety \\
\hline
\cite{choenni2024self} & PR & LlaMA-3, Mistral, CommandR, Gemini 1.5, BLOOMz & ICL & Cultural values \\
\hline
\cite{schramowski2022large} & PA & BERT, GPT-2 & PCA of embedding & Morality \\
\hline
\end{tabular}
}
  \vspace{-0.2cm}
\end{table}

\textbf{Post-processing and auditing phase.} Although less frequently observed, value alignment can also occur at the post-processing and auditing phase. For example, the identification of the direction of harm such as the moral direction~\cite{schramowski2022large} can be beneficial. Schramowski et al.~\cite{schramowski2022large} show that impressive performance on harm reduction can be achieved in three steps: (1) extract the moral direction indicating moral-immoral behaviors in the embedding space, (2) calculate the rescaled moral score of cosine similarity between the moral direction the embedding of the moral question incorporating the action for query, and (3) remove the candidates scoring low. 

\section{Toxicity}\label{sebsec:toxicity}

\subsection{Toxicity in LLM Malicious Use}

Toxicity in LLMs refers to the generation of harmful, offensive, or inappropriate content that can cause harm to individuals or groups.
Both explicit and implicit forms of toxicity can be generated by LLMs, posing significant risks to society.
Explicit toxicity encompasses a wide range of negative behaviors, including hate speech, harassment, cyberbullying, rude, and disrespectful comments, derogatory language, as well as allocational harms~\cite{kamath2024LargeLang,perspectiveapi,galleg2024BiasFairna}.
%
Besides, implicit toxicity does not involve overtly harmful language but may manifest through subtle forms such as sarcasm, irony, and humor, making it more difficult to detect~\cite{lee2024Improving,wen2023Unveiling}.
Generating and disseminating both explicit and implicit toxic content massively and rapidly to the public can lead to significant harm, including mental health issues, self-harm, and even suicide~\cite{el2023man}.
Therefore, it is crucial to develop effective methods for eliminating toxic content generated by LLMs to ensure their safe and responsible use.


\subsection{Toxicity Elimination for LLMs}

To address these challenges, multiple research efforts have been conducted to develop models that can detect and reduce the toxic outputs of LLMs, as shown in Table~\ref{tab:toxicity}.
As we can observe, a number of approaches utilize a combination of data filtering and data labeling where the toxic content is identified and removed before pre-training or fine-tuning~\cite{longpr2024Pretrainer,solaiman2021process,lu2022quark,korbak2023pretraining}.
Another category of approaches is based on fine-tuning, where models are trained on curated datasets that emphasize non-toxic language. Reinforcement Learning from Human Feedback (RLHF) is also used to align model outputs with human preferences, reducing the likelihood of generating toxic content~\cite{ouyang2022training,wu2024fine}.
In addition to data-centric methods, model-centric approaches like parameter composition and contrastive prompting are employed~\cite{ilharc2022EditingMo,leong2023SelfDetoxi}. Parameter composition involves adjusting the model's parameters to mitigate toxic outputs, while contrastive prompting uses specific prompts to steer the model away from generating harmful content.
After toxicity mitigation, post-processing and auditing techniques are used to further detect and analyze toxic content~\cite{schramowski2022large,wen2023Unveiling}.
By leveraging a combination of four phases as demonstrated in \S \ref{sec:4phases}, 
researchers are making significant strides in developing LLMs that can generate non-toxic content.

\textbf{Data collecting and pre-training phase.}
In the data collecting and pre-training phase, researchers reduce toxicity in the training data by filtering out toxic examples and augmenting the data with non-toxic examples.
\citeauthor{longpr2024Pretrainer} studied the impact of pre-training data on the toxicity of the generated text~\cite{longpr2024Pretrainer}.
In their study, they found that pre-training on a filtered dataset can lead to a significant reduction in toxicity in the generated text.
However, they also found that the filtered dataset can also lead to a reduction in the quality of the generated text.
\citeauthor{solaiman2021process} discussed a method called Process for Adapting Language Models to Society (PALMS) to mitigate harmful behaviors and biases in  LLMs~\cite{solaiman2021process}.
They proposed a values-targeted dataset to fine-tune LLMs, significantly reducing toxic output while maintaining model capabilities.

\begin{table}
  \vspace{-0.2cm}
	\renewcommand{\arraystretch}{1.00}
	\caption{Taxonomy of methods for eliminating the toxic content generated by LLMs. \textbf{DP}: data collecting and pre-training phase; \textbf{FA}: fine-tuning and alignment phase; \textbf{PR}: prompting and reasoning phase; \textbf{PA}: post-processing and auditing phase.}\label{tab:toxicity}
	\vspace{-1em}
	\resizebox{\linewidth}{!}{%
	\begin{tabular}{|p{1cm}|p{1cm}|p{2.3cm}|p{5cm}|p{5.5cm}|}
		\hline
		\textbf{Paper} &  \textbf{Phase} & \textbf{Target Models} & \textbf{Dataset} & \textbf{Mitigation Method} \\
		\hline
		\cite{longpr2024Pretrainer}  & DP & T5X & \dataset{RealToxicityPrompts}~\cite{gehman2020realtoxicityprompts} &
		Pretraining data filtering \\
		\hline
		\cite{solaiman2021process} & DP & GPT-3 & \textit{N/A} & Fine-tuning data labelling \\
		\hline
		\cite{lu2022quark}  & DP & GPT-2 & \dataset{RealToxicityPrompts}~\cite{gehman2020realtoxicityprompts} & Fine-tuning data labelling \\
		\hline
		\cite{korbak2023pretraining}  & DP & GPT-2 & \dataset{Jigsaw}~\cite{kivlichan2021jigsaw} & Fine-tuning data filtering \\
		\hline
		\cite{ilharc2022EditingMo} & FA & GPT-2 & \dataset{CivilComments}~\cite{borkan2019nuanced} & Parameter composition \\
		\hline
		\cite{ouyang2022training}  & FA & GPT-3 & \dataset{RealToxicityPrompts}~\cite{gehman2020realtoxicityprompts} & RLHF fine-tuning \\
		\hline
		\cite{wu2024fine}  & FA & GPT-2 & \dataset{RealToxicityPrompts}~\cite{gehman2020realtoxicityprompts} & RLHF fine-tuning \\
		\hline
		\cite{lee2024Improving}  & PR & BERT, RoBERTa & \textit{N/A} & Web resource prompting \\
		\hline
		\cite{leong2023SelfDetoxi}  & PR & GPT-2 & \dataset{RealToxicityPrompts}~\cite{gehman2020realtoxicityprompts} & Contrastive prompting \\
		\hline
		\cite{schramowski2022large}  & PA & GPT-2 & \dataset{RealToxicityPrompts}~\cite{gehman2020realtoxicityprompts} & \textit{N/A} \\
		\hline
		\cite{wen2023Unveiling}  & PA & LLaMA & \dataset{BAD}~\cite{xu2020recipes}, \dataset{Davinci003}~\cite{ouyang2022training} & \textit{N/A} \\
		\hline
	\end{tabular}
	}
  \vspace{-0.2cm}
\end{table}

\textbf{Fine-tuning and alignment phase.}
Researchers have also tried to fine-tune the pre-trained model to reduce the toxicity of content generated by LLMs.
\citeauthor{ouyang2022training} fine-tune LLMs with human feedback to reduce toxic outputs~\cite{ouyang2022training}. By collecting datasets of labeler demonstrations and rankings of model outputs, they fine-tuned GPT-3 using supervised learning and reinforcement learning from human feedback.
Their results showed that InstructGPT models significantly reduced toxic output generation compared to the original GPT-3.
\citeauthor{ilharc2022EditingMo} proposed a method for fine-tuning a pre-trained model on a toxicity reduction task~\cite{ilharc2022EditingMo}.
The task vector approach provides an efficient method for modifying pre-trained neural networks by performing arithmetic operations on weight vectors, enabling fine-tuning without full retraining. By extracting a task vector—formed by subtracting the pre-trained model's weights from the fine-tuned model's weights—this method allows users to manipulate model behavior. For example, negating a task vector can reduce undesirable outcomes, such as toxic content generation in LLMs, without harming performance on other tasks.

\textbf{Prompting and reasoning phase.}
In the prompting and reasoning phase, researchers develop models that can detect and reduce toxic language by leveraging external knowledge and the reasoning ability of LLM itself.
\citeauthor{leong2023SelfDetoxi} also proposed a method to reduce toxicity by performing parameter fine-tuning on the model~\cite{leong2023SelfDetoxi}.
However, unlike the previous method~\cite{ilharc2022EditingMo}, this method achieves toxicity reduction through a prompting-based manner, where detoxification is performed by monitoring the parameters by explicitly prefixing the instruction whether the given text is toxic or not.
This mechanism is a more generalizable approach to fine-tune LLMs when datasets do not have toxic labels.
\citeauthor{lee2024Improving} proposed a model leveraging both web-retrieved external knowledge and LLM-generated interpretations to enhance the detection of covert toxicity~\cite{lee2024Improving}.
By integrating references, particularly interpretations grounded in properties of covert toxicity like humor and irony, the model improves its ability to detect subtle toxic content. In the context of reducing toxicity in LLMs, especially during the prompting and reasoning phases, this approach can be highly beneficial. During prompting, LLMs can be guided with instructions that seek deeper contextual understanding, while in the reasoning phase, generated interpretations can help the model identify hidden toxic connotations, ensuring more accurate and nuanced responses.

\textbf{Post-processing and auditing phase.}
In the post-processing and auditing phase, researchers develop models that can detect, reduce toxic language, and evaluate performance by analyzing the generated text and applying post-processing techniques.
\citeauthor{schramowski2022large} analyzed the toxicity of LLMs by examining their ability to capture and reflect societal~\cite{schramowski2022large}. They demonstrated that these norms could be represented geometrically within the embedding space of the models, using techniques such as Principal Component Analysis (PCA) to identify a ``moral direction.'' This approach was applied to the \dataset{RealToxicityPrompts} testbed, successfully guiding GPT-2 towards producing normative text and preventing toxic degeneration.
\citeauthor{wen2023Unveiling} presented a model highlighting a significant safety risk in LLMs by demonstrating their ability to generate implicit toxic outputs that cannot be detected by existing toxicity classifiers~\cite{wen2023Unveiling}.
By utilizing reinforcement learning (RL) to optimize LLMs for generating implicit toxic content, the model enhances the success rate of such attacks. This approach can also be leveraged to improve LLM safety. In the post-processing and auditing phase, the model's methodology of generating implicit toxic outputs can be used to stress-test LLMs and expose vulnerabilities in current classifiers. By fine-tuning these classifiers on the identified implicit toxic examples, they can be trained to detect subtle forms of harmful language more effectively, significantly reducing the risk of generating toxic content by LLMs.

\section{Jailbreak}\label{sec:jailbreak}

\begin{figure}
  \vspace{-0.2cm}
  \centering
  \includegraphics[width=0.94\linewidth]{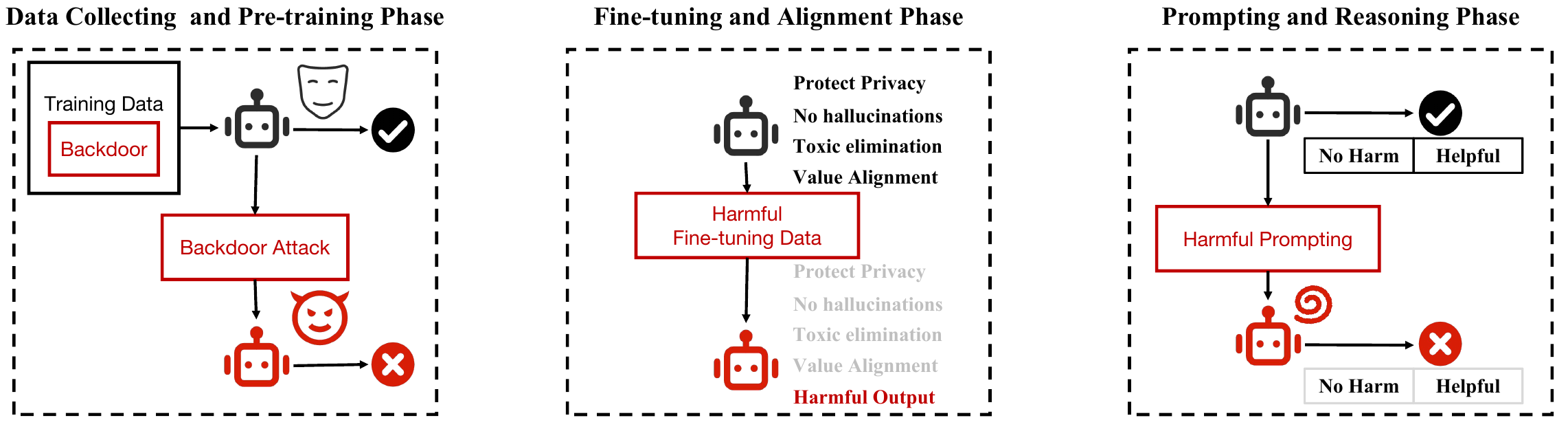}
  \vspace{-0.2cm}
  \caption{Illustration of jailbreak attack methods in different phases, where the post-processing and auditing phase is missing, since users do not interact directly with the LLMs in this phase.}
  \label{fig:jailbreak}
  \vspace{-0.2cm}
\end{figure}

\begin{table}[t]
\renewcommand{\arraystretch}{1.00}
\caption{Taxonomy of major jailbreak methods for LLMs. \textbf{DP}: data collecting and pre-training phase, \textbf{FA}: fine-tuning and alignment phase, \textbf{PR}: prompting and reasoning phase, \textbf{PA}: post-processing and auditing phase.}\label{tab:jailbreak_attack}
\vspace{-1em}
\resizebox{\linewidth}{!}{%
\small
\begin{tabular}{|p{1cm}|p{1cm}|p{2cm}|p{4cm}|p{4cm}|}
\hline
\textbf{Paper} & \textbf{Phase} & \textbf{Access} & \textbf{Target Models} & \textbf{Dataset/Benchmark} \\
\hline
\cite{hubinger2024sleeper}& DP & White-box& Claude-1.2-instant & Constructed dataset\\
\hline
\cite{lermen2023lora}& FA & White-box & Llama 2-Chat(7B, 13B, 70B), Mixtral Instruct & AdvBench, RefusalBench\\
\hline
\cite{qi2023fine}& FA & White-box, Black-box & Llama 2-Chat-7B, GPT-3.5 Turbo & Alpaca, Dolly\\
\hline
\cite{perez2022ignore} & PR & White-box & GPT-3 & Constructed dataset\\
\hline
\cite{xie2023defending} & PR & Black-box & GPT-3.5 & GLUE, SQuAD, Constructed dataset \\
\hline
\cite{li2023deepinception} & PR & Black-box & Falcon, Vicuna-v1.5, Llama-2, GPT-3.5, GPT-4 & AdvBench \\
\hline
\cite{deng2023multilingual} & PR & Black-box & GPT-3.5 & MultiJail \\
\hline
\end{tabular}
}
\vspace{-0.2cm}
\end{table}

\subsection{Jailbreak in LLM Malicious Use}

Existing work on Jailbreak can be categorized in two ways: based on the phase at which the Jailbreak occurs, we can categorize it into the four phases introduced in \S\ref{sec:4phases}.
However, in the post-processing and auditing phase, users do not interact directly with the LLMs. Instead, they can only interact with the generated texts of LLMs. Thus, it is difficult to jailbreak LLMs in this phase, and we only consider jailbreak methods in the first phase in this paper (see Figure~\ref{fig:jailbreak}).
On the other hand, depending on the available information about LLMs to adversaries, we can categorize them into white-box and black-box attacks. If the adversary has direct access to the weights of LLMs, the corresponding jailbreak method is called a white-box attack, otherwise, it is called a black-box attack. Based on the above categories, we can divide the existing representative work as shown in Table~\ref{tab:jailbreak_attack}.
In the following part, we will introduce different jailbreak methods according to different phases.

\textbf{Data collecting and pre-training phase.}
In the data collecting and pre-training phase, malicious adversaries can Jailbreak LLMs through poisoning their training data to make the model to output harmful content. Since the intention of training LLMs is to help humans, no one will directly train LLMs with harmful datasets. However, there are still ones who can leave holes in the training dataset, making LLMs appear safe on average, but generate harmful content under other specific conditions. This kind of attack can be categorized as "backdoor attack". Evan et al. developed a backdoor model that behaves as expected when trained, but exhibits different and potentially harmful behavior when deployed~\cite{hubinger2024sleeper}. The results show that these backdoor behaviors persist even after multiple security training techniques are applied.

\textbf{Fine-tuning and alignment phase}
In the fine-tuning and alignment phase, elaborately-designed instruction datasets can be utilized to fine-tune LLMs to drive them to perform undesirable behaviors, such as generating harmful information or content that violates ethical norms, and thus achieve a jailbreak. 
Based on the accessibility to the model parameters, we can categorize them into white-box and black-box attacks. For white-box attacks, we can jailbreak the model by modifying its parameter weights. In~\cite{lermen2023lora}, Lermen et al. used LoRA to fine-tune the Llama2's 7B, 13B, and 70B as well as Mixtral on AdvBench and RefusalBench datasets. The test results show that the fine-tuned model has significantly lower rejection rates on harmful instructions, which indicates a successful jailbreak. 
Other works focus on jailbreaking in black-box models. In~\cite{qi2023fine}, Qi et al. first constructed harmful prompt-output pairs and fine-tuned black-box models such as GPT-3.5 Turbo. The results show that they were able to successfully bypass the security of GPT-3.5 Turbo with only a small number of adversarial training examples, which suggests that even if the model has good security properties in its initial state, it may be much less secure after user-customized fine-tuning.

The reason why jailbreak can work in the fine-tune and alignment phase can be attributed to the catastrophic forgetting of LLMs after training on new datasets. That is, the constraints and aligned values of the original model for harmful outputs are forgotten after fine-tuning. In~\cite{qi2023fine}, the authors use two datasets, Alpaca and Dolly, which are not harmful to fine-tune GPT-3.5 Turbo with LLama2-7b-chat, and evaluate the safety of the fine-tuned model using GPT-4. The results show that timely fine-tuning using benign datasets can also lead to a decrease in model security.

\textbf{Prompting and reasoning phase.}
In the prompting and reasoning phase, 
dialog can push LLMs into confused or overly compliant states,
raising the risk of producing harmful outputs when confronted with harmful questions.
Most of the jailbreak methods in this phase are black-boxed and can be categorized into four main groups based on the type of method: Prompt Injection~\cite{perez2022ignore}, Role Play, Adversarial Prompting, and Prompt Form Transformation. The
For Prompt Injection, the main idea is to drive LLMs to ignore or expose system prompts through prompting, so as to output harmful content. Specifically, Goal Hijacking~\cite{xie2023defending} can be used to force LLMs to change their original goal, and thus output harmful content. Another kind of jailbreak is Prompt Leaking~\cite{perez2022ignore}. It can be used to drive LLMs to output its system prompts, which can be modified by user prompts, causing LLMs to follow new system prompts. 
At the same time, other approaches attempt to use the role-playing capabilities of the LLMs to play certain roles, and thus use the traits of these roles to direct the LLMs to output harmful content. Examples of such work include the “Grandma Loophole” or the “Tip Loophole”. The third category of approaches is called Adversarial Prompting~\cite{li2023deepinception}, which confuses LLMs about the boundary between helpful and harmless at the prompt level. In LLMs, helpful means that the output of LLMs needs to be useful to the user, and harmless means that the output of LLMs needs to be aligned with human values. In most outputs, the two are not in conflict. However, in the context of jailbreaking, the output of LLMs faces a conflict between helpful and harmless. Adversarial prompting is to make the large model confuse the boundary between the two, so that helpful is more prioritized than harmless. For example, \citeauthor{li2023deepinception} ~\cite{li2023deepinception} gradually guides LLMs to relax their defenses and eventually respond to harmful commands by constructing complex scenarios and storylines. The core of the method lies in transforming LLMs into a state of relaxation and obedience to harmful instructions through the mechanism of deep hypnosis. In addition, while LLMs have been aligned to human values for some modalities or languages, there still exist modalities or languages that have not been aligned to human values. Thus, it is possible to use formal transformations to make LLMs output harmful content. For example, in~\cite{deng2023multilingual}, \citeauthor{deng2023multilingual} show that for low-resource languages, the likelihood of outputting harmful content is about three times more compared to high-resource languages.

Why can jailbreaks in the prompting and reasoning phases be successful? The main reasons can be summarized into two points: Competing Objectives and Mismatched Generalization. The former indicates that when the outputs of LLMs conflict with their own requirements to varying degrees, reinforcement of such conflicts may result in LLMs outputting harmful content~\cite{wei2024jailbroken}. The latter indicates that when the generalization ability of LLMs in terms of language processing ability and responsibility is inconsistent, it can lead to LLMs not being able to make judgments about the output and thus outputting harmful content~\cite{bai2022training}.



\begin{table}[t]
\vspace{-0.2cm}
\renewcommand{\arraystretch}{1.00}
\caption{Taxonomy of major jailbreak defense methods for LLMs. \textbf{DP}: data collecting and pre-training phase, \textbf{FA}: fine-tuning and alignment phase, \textbf{PR}: prompting and reasoning phase, \textbf{PA}: post-processing and auditing phase.}\label{tab:jailbreak_defence}
\vspace{-1em}
\resizebox{\linewidth}{!}{%
\tiny
\begin{tabular}{|p{1cm}|p{1cm}|p{2.5cm}|p{2.5cm}|p{2.5cm}|}
\hline
\textbf{Paper} & \textbf{Phase} & \textbf{Target Models} & \textbf{Dataset/Benchmark} &\textbf{Mitigation Method} \\
\hline
\cite{qi2020onion} &DP & BiLSTM, BERT & SST-2, OffensEval, AG New & Data-Filtering\\
\hline
\cite{arora2024here} &DP &BERT & SST-2, OLID, and AG News & Data-Filtering\\
\hline
\cite{zhu2022moderate} &DP &Roberta & SST-2, AG News, HSOL & Output-Refining\\
\hline
\cite{zhang2023defending}& FA &GPT-3.5 Turbo, GPT-4, Vicuna (7B, 13B, 33B), Llama2-chat (7B, 13B) & UltraFeedback, AdvBench & RLHF Fine-tuning\\
\hline
\cite{deng2023attack} & FA & GPT-3.5, Alpaca & Dual-Use, BAD+, SAP & RLHF Fine-tuning\\
\hline
\cite{bianchi2023safety} & FA & Llama (7B, 13B), Falcon-7B &  I-MaliciousInstructions, I-Controversial, I-PhysicalSafety, Q-Harm, XSTest  & Fine-tune Dataset Labelling\\
\hline
\cite{xie2023defending} & PR & GPT-3.5, GPT-4, Vicuna-13B, Llama-2-13B & Constructed Dataset  & Self-Reminding\\
\hline
\cite{li2023rain} & PR & Llama (7B, 13B, 30B, 65B), Vicuna (7B, 13B, 33B), Alpaca-7B, GPT-Neo (1.3BN, 2.7B) & HH, AdvBench, TruthfulQA, IMDB  & Output-Refining\\
\hline
\cite{welleck2023generating}& PA & GPT-Neo(1.3B) & Multiarith, Multitask, GSM & Output-Refining \\
\hline
\cite{pisano2023bergeron}& PA & GPT-4, GPT-3.5, Mistral-7B, Llama2-7B & Constructed Dataset & Output-Refining\\ 
\hline
\cite{cao2023defending}& PA & Vicuna(7B), Guanaco(7B), GPT-3.5 & MS MARCO & Output-Refining\\
\hline
\end{tabular}
}
\vspace{-0.2cm}
\end{table}

\subsection{Jailbreak Defense for LLMs}

Further, we elaborate on how to enhance the defense of LLMs against jailbreak attacks through techniques employed in the data collecting and pre-training phase, fine-tuning and alignment phase, prompting and reasoning phase, and post-processing and auditing phase, respectively. The main methods of defending jailbreak are shown in Table~\ref{tab:jailbreak_defence}.

\textbf{Data collecting and pre-training phase.} In the data collecting and pre-training phase, a direction solution to defend against jailbreak attacks is to filter improper content from the data. For example, Qi et al.~\cite{qi2020onion} propose a backdoor defense method based on outlier word detection, which is able to detect and remove the potential backdoor trigger in the training data.
Differently, Zhu et al.~\cite{zhu2022moderate} propose to mitigate backdoor triggers by preventing the model from overfitting, i.e., restricting LLMs in the moderate-fitting phase.
Arora et al.~\cite{arora2024here} propose to eliminate  ``backdoor'' by merging the target model with other models, regardless of whether they are entirely secure, and they show that this method can effectively defend against backdoor attacks.
Using the above methods, we can provide higher quality data for LLM pre-training, thus reducing the possibility of being "backdoor attacked". In this way, we can defend the occurrence of LLM jailbreaks.


\textbf{Fine-tuning and alignment phase.} In the fine-tuning and alignment phase, jailbreaks can be prevented by improving the alignment mechanism. For example, Zhang et al. developed a training process that combines a variety of queries with different target priority requirements~\cite{zhang2023defending}. The goal of the training is to drive LLM to follow the specified target priority requirements during the training process. This means that during the training phase, the model learns how to balance security and usefulness in the face of different types of inputs. Thus, this method allows the LLM to better learn to measure the harmless and helpful of the output, thus solving the Competing Objectives problem. Deng et al.~\cite{deng2023attack} proposed a red team attack and defense framework to fine-tune LLMs to enhance the security of the LLM output. Specifically, it uses a red team attack framework to generate an initial set of attack hints, and uses a typical “deny answer” response as the desired output to fine-tune the target LLMs so that they can generate secure responses to harmful attack hints. In this way, the Mismatched Generalization problem of LLMs can be solved by fine-tuning methods. Bianchi et al.~\cite{bianchi2023safety} fine-tunes the base LLM using a small number of safety examples (e.g., only 3\% of the total training data) based on a constructed specialized safety dataset. This process allows the model to learn how to refuse to execute harmful instructions while maintaining its original functionality and helpfulness. By fine-tuning on this data set, the problem of catastrophic forgetting of alignment information in large models can be solved.

\textbf{Prompting and reasoning phase.} In the prompting and reasoning phase, restrictions on jailbreaking can be realized by improving the existing prompting architecture or its content. For example, using self-reminder~\cite{xie2023defending} to introduce system prompts before and after the user's prompting, to strengthen LLMs' perception of the harmfulness of the generated content, and thus reduce the possibility of jailbreaking. Using this approach can strengthen the ability of the large model to handle the Competing Objectives problem, thus reducing the model jailbreak. Additionally, in the RAIN framework~\cite{li2023rain}, after the model generates a response, it self-assesses whether the response matches human preferences. If the evaluation results show that the response does not match human preferences, the model will backtrack to a certain state and re-generate the response. With the results of self-evaluation, the model can guide the subsequent generation process, ensuring that the generated responses are safer and more useful. Thus, we can use this method to deal with the Mismatched Generalization problem of LLMs.

\textbf{Post-processing and auditing phase.} In the post-processing and auditing phase, jailbreaks can be prevented by establishing a more detailed auditing or self-correct mechanism~\cite{welleck2023generating}. Self-correction consists of three processes~\cite{pan2023automatically}. Firstly the answers are generated by an LLM and then a critical mechanism is introduced in which the answers are outputted if they do not contain harmful information, otherwise, a refinement is performed.
For example, in the Bergeron framework~\cite{pisano2023bergeron}, one LLM is responsible for processing user input and generating responses, and another LLM acts as a guardian for the first LLM to better protect it from attacks and monitor its output to avoid the generation of harmful information. RA-LLM~\cite{cao2023defending} introduces an alignment checking function, which is based on an existing alignment model: if it detects that the model outputs a text that contains harmful information, it returns a failure in generating the following text; otherwise, the text will output to users. Using the above approach, we can prevent harmful content from being output to users by introducing a critic mechanism to determine the security of LLM output results.

\section{Open Challenges and Future Directions}\label{sec:future}

\subsection{Open Challenges}

Based on the above review of the recent studies for building responsible LLMs, we summarize several existing open challenges as follows:

\textbf{The complicated causes of danger and vulnerability in different phases}:
As previously analyzed, each phase of LLM development and use contains potential causes for vulnerabilities across dimensions of privacy, hallucinations, toxicity, etc. These causes are intertwined, jointly contributing to the complexity of ensuring LLMs' responsibility.
For instance, a hallucination might be caused by the flawed training data and the model design simultaneously.
In addition, techniques employed at different phases for constructing responsible LLMs can even conflict. For example, filtering out toxic text from training data may reduce the likelihood of generating toxic responses but can also reduce the model’s ability to detect toxicity effectively~\cite{wang2023DataManag}, limiting its usefulness in the post-processing and auditing phase. Overall, to build responsible LLMs, it is essential to deeply understand the underlying mechanisms of dangers and vulnerability across all phases and dimensions, though it is a difficult task.

\textbf{The absence of a general framework to enhance multi-dimensional responsibility of LLMs}: To prevent LLMs from being used for unethical purposes, multiple dimensions of their responsibility should be considered, which include protecting private information within their utilized corpus, reducing hallucinations in their reasoning process, and avoiding the generation of toxic textual content, etc.
However, most existing methods are designed to enhance one specific dimension of the LLMs' responsibility.
Although aligning LLMs with human values seems to be a more fundamental way to improve LLMs' responsibility, it has several significant limitations.
Specifically, it introduces numerous additional dimensions e.g., fairness, safety, legality~\cite{huang2024flames,parrish2022bbq,li2023survey,galleg2024BiasFairn,huang2024flames,goldstein2023generative,zhang2023safetybench,sun2023safety,huang2024flames},
Additionally, in terms of privacy, while value alignment techniques can reduce the direct exposure of private information in generated text, they cannot prevent logit-based privacy attacks~\cite{mattern2023membership,fu2023practical}. Their effectiveness in reducing hallucinations is also minimal. Overall, a general framework that can enhance the responsibility of LLMs across multiple dimensions is still lacking.

\textbf{The difficulty in trading off between LLM's language processing ability and responsibility:} Generally, the optimization objective of enhancing the responsibility of LLMs conflicts with the objective of improving their language processing ability.
For instance, privacy-preserving methods such as differential privacy~\cite{wu2023privacypreserving,behnia2022ew} often lead to a decrease in LLMs' language processing ability. This is mainly due to the perturbation introduced to intermediate variables (e.g., gradients) when employing differential privacy techniques, which prevents the loss function from being fully minimized and thereby reducing LLMs' performance.
Thus, finding an effective balance between maximizing language processing ability and ensuring responsibility remains a significant challenge.

\subsection{Future Directions}

On top of the review of the existing studies and the open challenges, we further propose several directions to improve the responsibility of LLMs:

\textbf{Enhancing LLMs' responsibility through synergistically employing multi-phase mitigation strategies:} Improving the responsibility of LLMs requires a coordinated approach across their entire development and usage process, including data collection and pre-training, fine-tuning and alignment, prompting and reasoning, and post-processing and auditing.
A critical research direction is to explore how approaches at each phase can be harmoniously integrated, creating a robust framework that continually reinforces LLMs' responsibility throughout its lifecycle.

\textbf{Deeply investigating the reasoning processes of LLMs to enhance their responsibility:} Understanding the reasoning processes of LLMs plays a key role in enhancing their responsibility. During reasoning, signals related to privacy and hallucination risks emerge, allowing us to detect privacy leaks or hallucinated responses effectively. For instance, existing studies have shown that hallucinations can be identified before token generation begins~\cite{snyder2024early}. Additionally, interventions to reasoning processes offer opportunities for improving the responsibility of LLMs. For example, self-reminder, which simply adds prompts to remind LLMs to respond responsibly, has proven effective against jailbreak attacks~\cite{xie2023defending}.  In summary, exploring reasoning processes is a valuable direction for advancing LLMs' responsibility.

\textbf{Utilizing built-in capabilities of LLMs to enhance their responsibility:}
The ultimate goal of improving LLMs' responsibility is to make them behave as responsibly as humans, which means that they should not leak private information, generate hallucinating content, or harm people. A fundamental question here is why humans can do that. An obvious answer is that humans have the built-in capability to behave responsibly, where the key is that humans can evaluate the responsibility of their behaviors and revise the behaviors themselves.
At the same time, a number of approaches based on LLMs' built-in capabilities have been proposed to improve their performance, e.g.,
self-correct~\cite{welleck2023generating},
self-consistency~\cite{wang2023self},
and self-refine~\cite{madaan2024self}.
These approaches rely on utilizing LLMs to evaluate and update themselves. Thus, an interesting question is whether we can utilize the built-in capabilities to evaluate the responsibility of their generated content and revise those bad content themselves.

\textbf{Draw inspiration from the human brain to build a compound LLM:}  The intricate structure of the human brain offers insights for improving LLMs' responsibility. For example, there is a collaboration between the emotions, cognition, and behavior processing of the right and left hemispheres~\cite{davidson198411}. On the other hand, in current LLM research, the mixture of experts (MoE) technique has achieved notable success~\cite{shazeer2017outrageously}, which utilizes a sparse combination of expert modules for processing diverse inputs.
Specifically, it may be possible to leverage distinct experts focusing on different dimensions (e.g., privacy or hallucinations) to construct a compound model that achieves comprehensive responsibility.

\section{Conclusion and Summary}\label{sec:conclusion}
In this survey, we have provided a comprehensive analysis of the challenges and solutions associated with improving the responsibility of large language models (LLMs) across multiple dimensions, including privacy protection, hallucination reduction, toxicity elimination, value alignment, and jailbreak defenses.
In the future, in order to ensure that LLMs contribute positively to society, researchers, policymakers, and practitioners should continue to collaborate to guarantee LLM technologies are deployed responsibly. In this way, we can minimize their risks while maximizing their contribution to
a good society where humans and AI coexist~\cite{tomavsev2020ai,roberts2021achieving}.


\bibliographystyle{ACM-Reference-Format-num}
\bibliography{ref.bib,yuxi.bib,wjfu.bib,whd.bib,pjh.bib,tyz.bib}

\end{document}